\pdfoutput=1
\documentclass[11pt]{article}

\usepackage[preprint]{acl}

\usepackage{authblk}

\usepackage{times}
\usepackage{soul}
\usepackage{url}
\usepackage{natbib}
\usepackage{multicol,multirow}
\usepackage[utf8]{inputenc}
\usepackage{caption}
\usepackage{graphicx}
\usepackage{amsmath}
\usepackage{amsthm}
\usepackage{booktabs}
\usepackage{algorithm}
\usepackage{algorithmic}
\usepackage[switch]{lineno}
\usepackage{xcolor}         % colors
\usepackage{color, colortbl}
\usepackage{enumitem}
\usepackage{subcaption}
\usepackage{hyperref}
\usepackage{comment}
\usepackage{amsfonts} 
\newcommand{\OURS}{\textsc{Step-back Profiling}}
\newcommand{\july}[1]{\textcolor{black}{#1}}

\title{\OURS{}: Distilling User History for Personalized Scientific Writing}

\author[1]{\textbf{Xiangru Tang\thanks{Equal Contribution.}}}
\author[2]{\textbf{Xingyao Zhang$^*$}}
\author[1]{\textbf{Yanjun Shao$^*$}}
\author[2]{\textbf{Jie Wu}}
\author[1]{\textbf{Yilun Zhao}}
\author[1]{\\ \textbf{Arman Cohan}}
\author[2]{\textbf{Ming Gong}}
\author[2]{\textbf{Dongmei Zhang}}
\author[1]{\textbf{Mark Gerstein}}

\affil[1]{Yale University}
\affil[2]{Search \& Distribution, Microsoft}

\affil[ ]{\texttt{xiangru.tang@yale.edu}}

\begin{document}

\maketitle

\begin{abstract}
Large language models (LLM) excel at a variety of natural language processing tasks, yet they struggle to generate personalized content for individuals, particularly in real-world scenarios like scientific writing.
Addressing this challenge, we introduce \OURS{} to personalize LLMs by distilling user history into concise profiles, including essential traits and preferences of users. To conduct the experiments, we construct a Personalized Scientific Writing (PSW) dataset to study multi-user personalization.
PSW requires the models to write scientific papers given specialized author groups with diverse academic backgrounds. 
As for the results, we demonstrate the effectiveness of capturing user characteristics via \OURS{} for collaborative writing. 
Moreover, our approach outperforms the baselines by up to 3.6 points on the general personalization benchmark (LaMP), including 7 personalization LLM tasks.
Our ablation studies validate the contributions of different components in our method and provide insights into our task definition. Our dataset and code are available at \url{https://github.com/gersteinlab/step-back-profiling}.
%Our dataset and code will be available upon acceptance.
\end{abstract}

\section{Introduction}

\vspace{-3pt}
% \setlength{\epigraphwidth}{0.95\columnwidth}
% %\renewcommand{\epigraphflush}{center}
% %\renewcommand{\epigraphsize}{\footnotesize}
% \epigraph{``History is a guide to navigation in perilous times. History is who we are and why we are the way we are.''}
% {\textit{David McCullough~\citep{mccullough20051776}}}

Recently, Large Language Models (LLMs) have made significant progress in natural language understanding and generation
%, demonstrating human-parity performance on a wide range of tasks 
\citep{wei2022emergent, zhang2023igniting, openai2023gpt4, qin2023toolllm}. 
%Moreover, the advent of LLM-driven language agents has revolutionized a myriad of user-facing applications, marking a game-changing breakthrough in the general AI capacity \citep{zhou2023agents, zhang2023igniting, shinn2023reflexion, qin2023toolllm, qiao2024autoact}.
Concurrently, integrating LLMs with personalization paradigms has paved the way for a vast frontier in improving user-centric services and applications \citep{salemi2023lamp, chen2023large, zhiyuli2023bookgpt}, as they provide a deeper understanding of users' accurate demands and interests than abstract vector-based information representations. By learning to characterize and emulate user-specific language patterns, personalized LLMs can enable more engaging and valuable interactions in domains such as dialogue \citep{wang2019persuasion, zhang2019dialogpt,characterai2022}, recommendation \citep{zhiyuli2023bookgpt, wang2023recmind}, \july{role-playing \citep{shao2023character,jiang2023personallm,tang2024medagents}} and content creation \citep{wei2022creater,cao2023comprehensive}.

\begin{figure}[t]
\centering
\includegraphics[width=\linewidth]{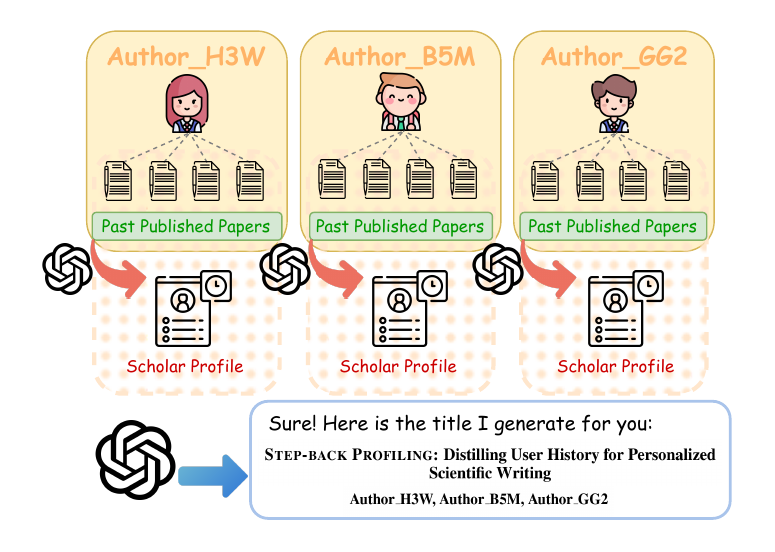}
\caption{\textbf{Overview of the \OURS{}.} \july{\OURS{} creates compact profile representations for each user and augments LLM generation with personalized information.}}
%It applies the abstraction function to the history of users and generates personalized output through an enhanced retrieval-augmented language model.
\label{fig:1}
\vspace{-.6cm}
\end{figure}

Prior work on personalizing language models \citep{salemi2023lamp, tan2023user, zhang2023memory, chen2023large, zhiyuli2023bookgpt} has shown promise, but primarily focused on learning user representations in a single-user context.
%For example, the LaMP benchmark \citep{salemi2023lamp} evaluates personalization given a single target user's historical interactions on tasks like citation prediction and product review generation. 
However, many real-world applications involve multiple users collaborating on a shared task, such as team-authored scientific papers. 
Another practical challenge for LLM personalization is scaling to extensive user histories while respecting context length limits \citep{shi2023large, liu2024lost,zhang2024cogenesis}. Directly conditioning on raw personal histories quickly becomes infeasible as user data grows. Prior methods mostly use uncompressed history for personalization \citep{salemi2023lamp}, which restricts the amount of user-specific information the model can utilize. 
%This limits knowledge-intensive applications like scientific writing, where relevant information may be dilated across many documents.

\begin{figure}[t]
\centering
    \vspace{-.35cm}
\includegraphics[width=\linewidth]{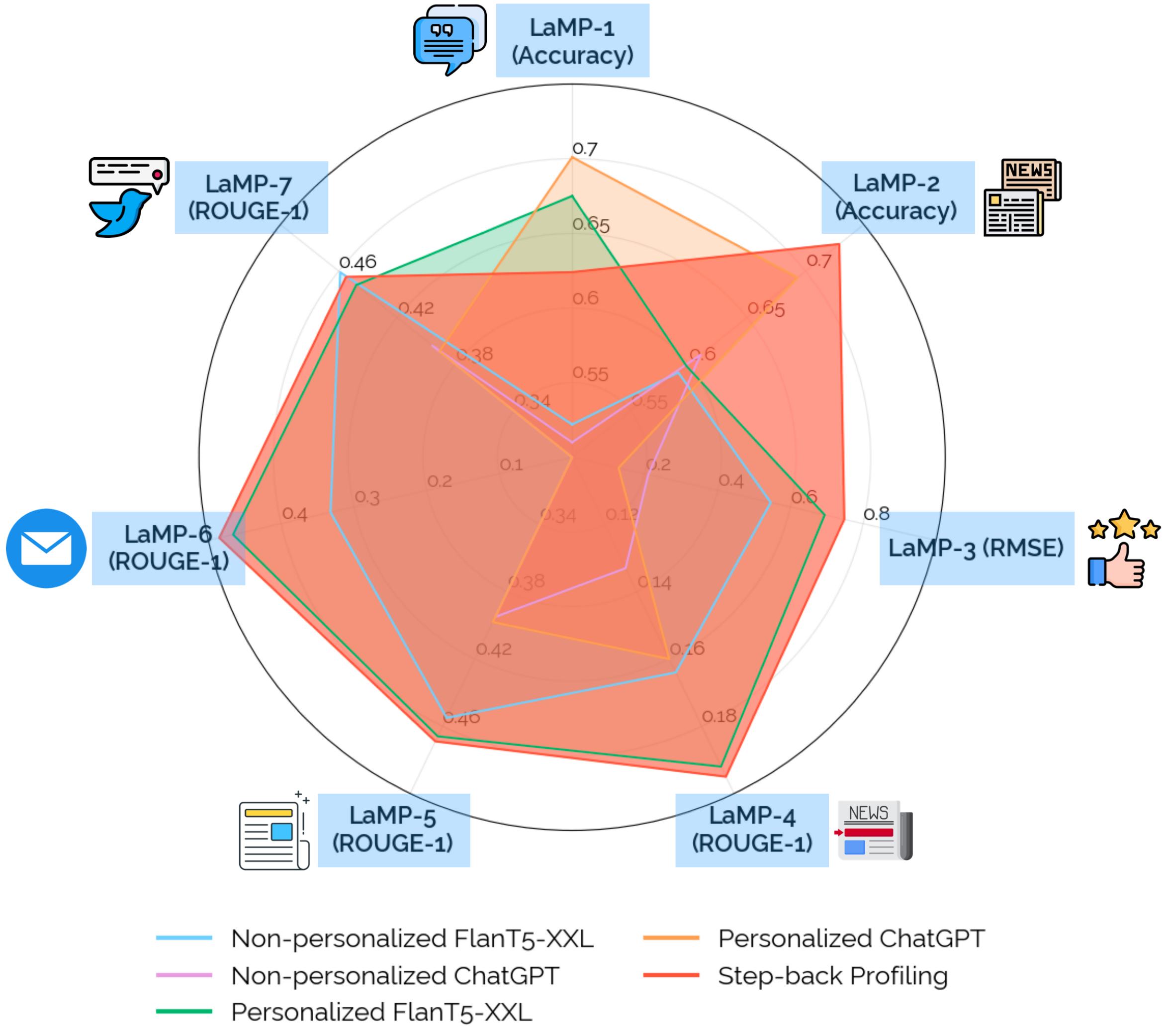}
\vspace{-.3cm}
\caption{\textbf{\OURS{} performance on the LaMP benchmark.} \july{We observe dominant performance of \OURS{} on most LaMP tasks.} Details of experimental setup can be found in Section \ref{sec:main_result}.}
\label{fig:2}
    \vspace{-.5cm}
\end{figure}

\july{In the realm of scientific publishing, we observe two distinct paradigms: papers authored by individuals and those resulting from team science collaborations. The latter category, exemplified by consortium science projects, can involve hundreds of authors working together \citep{wuchty2007increasing}. This shift towards large-scale collaborative research necessitates a deeper understanding of how to effectively model and generate content that reflects the collective expertise and writing styles of multiple authors. Our work aims to address this gap by developing methods to mimic and enhance the collaborative writing process in team science scenarios.}

As shown in Figure \ref{fig:1}, this work proposes a training-free LLM personalization framework that addresses these challenges through \OURS{}, 
%-- inspired by the ideas of gist memory \citep{lee2024human} and \textsc{Step-back Prompting} \citep{zheng2023take} for information compression and abstraction, 
we distill individual user histories into concise profile representations that capture high-level concepts and language traits. 
This enables efficient memory management and allows the model to focus on salient user characteristics, grounding personalized generation without excess computation or laborious data collection \citep{chen2023large}. 
%\OURS{} is a low-cost and easy-to-implement technique that operates directly on the pre-trained LLM without additional training. It can also complement parameter-efficient techniques of LLM finetuning \citep{hu2021lora, dettmers2024qlora, sheng2023s}. 
We show that \OURS{} improves performance over standard personalization methods (retrieval-based) in the LaMP\footnote{https://lamp-benchmark.github.io/}, as shown in Figure \ref{fig:2}.
Moreover, we introduce a Personalized Scientific Writing (PSW) dataset to study multi-user personalization. PSW contains research papers collaboratively written by expert teams, and each author's background publications are used to construct profiles. Modeling a group's collective expertise is crucial for this task, as different paper sections may reflect knowledge associated with particular authors. PSW thus poses a challenging and realistic testbed for multi-user personalization, requiring both abstractions of individual expertise and dynamic integration of diverse user traits throughout the collaborative writing process.

\july{By focusing on multi-author personalization, our work not only addresses a critical gap in current LLM research but also aligns with the evolving landscape of scientific collaboration. The PSW dataset and our \OURS{} approach provide a foundation for understanding and enhancing the complex dynamics of team-based scientific writing, potentially leading to more efficient and effective collaborative research practices.}

\begin{comment}
To summarize the contributions of this work:

\begin{enumerate}[topsep=0pt,parsep=1pt]
    \item A training-free \OURS{} approach that enables efficient and expressive personalization by abstracting user histories into trait-centric representations.
    \vspace{-.3cm}
    
    \item The Personalized Scientific Writing (PSW) dataset, a real-world benchmark for studying multi-user personalization with a novel task of collaborative expert writing.
        \vspace{-.3cm}
        
    \item A state-of-the-art performance of \OURS{} for single and multi-user personalization on diverse tasks in the LaMP and PSW benchmark without additional training.
\end{enumerate}
\end{comment}

\vspace{-.1cm}
\section{\OURS{}}

\vspace{-.1cm}

\subsection{Motivation}

Existing methods for personalizing language models struggle to effectively utilize user histories, particularly in the presence of extraneous details that can obscure the most pertinent information for a given task \citep{shi2023large, liu2024lost}. This challenge is magnified in multi-user scenarios, where models must efficiently extract and integrate knowledge from multiple users' histories. While retrieval-augmented methods, such as those employed in the LaMP benchmark \citep{salemi2023lamp}, have made progress in scaling to more extensive user histories, they still operate on raw user data containing relevant and irrelevant details. To address these limitations, we introduce a \OURS{} approach that distills a user's raw history into a concise representation focusing on 'gist' representations and preferences.
%drawing inspiration from the \textsc{Step-back Prompting} technique \citep{zheng2023take} and the \textsc{ReadAgent} framework \citep{lee2024human}. 
Our approach aims to enable more efficient and effective personalization across diverse single and multi-user scenarios by reasoning about higher-level traits instead of verbatim user history.
\vspace{-.2cm}

\subsection{Procedure}

Consider a set of $n$ users denoted by $\mathbb{U}$ = $\{ u_i \}_{i=1}^{n}$, where each user $u_i$ has a preference history $\mathbb{H}_i = \{(x_{ij}, y_{ij})\}_{j=1}^{m}$ consisting of $m$ input-output pairs. To effectively generate $P(y | x, \mathbb{H}_\mathbb{U})$ based on users' preference history, we create a set of user profiles $\mathbb{P}_\mathbb{U} = \{\mathbb{P}_{u_i}|u_i\in\mathbb{U}\}$ using \OURS{}. The complete procedure involves the following steps:
\vspace{-.3cm}

\paragraph{User Profile Gisting:} Each user's history is condensed into a short ``gist'' representation using an abstraction function $\operatorname{Gist}(\cdot)$: $\mathbb{P}_{u_i} = \operatorname{Gist}(\mathbb{H}_i)$. The ``gist'' captures the user's high-level traits and interests. \july{In \OURS{}, the $\operatorname{Gist}(\cdot)$ is also achieved using a language model. Details of the prompts can be found in Appendix \ref{sec:prompt_lamp} and \ref{sec:prompt_psw}.}

\vspace{-.3cm}

\paragraph{Multi-User Profile Concatenation:} Individual user profiles $\mathbb{P}_{u_1}$, $\mathbb{P}_{u_2}$, $\cdots$, $\mathbb{P}_{u_n}$ are concatenated to form a unified representation $\mathbb{P}_{\mathbb{U}} = [\mathbb{P}_{u_1}; \mathbb{P}_{u_2}; \cdots; \mathbb{P}_{u_n}]$, where $[\cdot; \cdot]$ is \july{an \textit{order-sensitive} function} combining the user profiles.

\vspace{-.3cm}

\paragraph{Retrieval-Augmented Generation (Optional):} Relevant snippets from user histories $\mathbb{H}_\mathbb{U}$ may be retrieved for input $x$ using a retrieval function $\operatorname{Retrieve}(\cdot)$. We have $\mathbb{R}_{i,k} = \operatorname{Retrieve}(x, \mathbb{H}_i, k)$, where $\mathbb{R}_{i,k}$ is a set of top-$k$ retrieved input-output snippets from user $u_i$'s history $\mathbb{H}_i$. The top-$k$ retrieved snippets $\mathbb{R}_k = \{\mathbb{R}_{i,k}\}_{i=1}^{N}$ can be concatenated with $x$ to form an augmented input $\tilde{x} = [x; \mathbb{R}_{1,k}; \mathbb{R}_{2,k}; \cdots; \mathbb{R}_{n,k}]$.

\vspace{-.3cm}

\paragraph{Personalized Output Generation:} The personalized language model generates an output $y = \operatorname{Generate}(\tilde{x}, \mathbb{P}_\mathbb{U})$ by conditioning on the augmented input $\tilde{x}$ (if retrieval is used) or the original input $x$, along with the concatenated user profile $\mathbb{P}_\mathbb{U}$. The generated output $y$ aligns with the user preferences captured by the \OURS{} while following the input $x$.

\section{The Personalized Scientific Writing (PSW) Benchmark}

\subsection{Motivation}

\july{The Personalized Scientific Writing (PSW) Benchmark is motivated by the evolving landscape of scientific research, where large-scale collaborations involving numerous authors across disciplines have become increasingly common \citep{wuchty2007increasing, wu2019large}. While existing benchmarks like LaMP \citep{salemi2023lamp} focus on single-user personalization, they fail to address the complex dynamics of multi-author scenarios prevalent in modern scientific writing. PSW aims to bridge this gap by providing a framework for evaluating personalized language models in collaborative contexts, challenging them to capture and reproduce the intricate interplay of individual authors' styles, expertise, and roles within a team \citep{fortunato2018science}.}

\july{By extending personalization to multi-user scenarios, PSW reflects real-world applications in interdisciplinary research, where integrating diverse knowledge and writing styles is crucial. The benchmark assesses how well language models can synthesize collective intelligence in writing tasks, a critical factor in enhancing the quality and impact of research outputs \citep{lee2021scientific}. Through PSW, we aim to drive advancements in personalized language modeling techniques that can handle multiple user profiles simultaneously, potentially contributing to more efficient and impactful research practices in the era of team science and artificial intelligence-assisted discovery \citep{king2009automation}.}

%We have extended the LaMP benchmark, introduced by \cite{salemi2023lamp}, to evaluate multi-user scenarios. Our PSW benchmark includes four tasks, and we outline the data collection process for each task.
\vspace{-.1cm}

\subsection{Problem Formulation}

Personalized language models aim to generate outputs that follow a given input and align with the users' styles, preferences, and expertise. In multi-author collaborative writing, 
%the Personalized Writing Styles (PSW) benchmark provides a framework for evaluating such models.
each data entry in the PSW benchmark consists of four key components:
(1) An input sequence $x$ serves as the model's input;
(2) A target output $y$ that the model is expected to generate;
(3) A set of user histories $\mathbb{H}_\mathbb{U} = \{\mathbb{H}_{u_i}\}_{i=1}^{l}$, where $l$ is the number of collaborating authors, and each entry $\mathbb{H}_{u_i}$ contains historical input-output pairs for user $u_i$;
(4) A set of author roles $\mathbb{C} = \{c_i\}_{i=1}^{l}$, each representing the role of the corresponding author $u_i$ in the collaborative writing process.

A personalized language model aims to generate an output $y$ that aligns with the conditional probability distribution $P(y | x, \mathbb{H}_\mathbb{U}, \mathbb{C})$. This means the model should produce an output that follows the input $x$ and the collaborating authors' writing styles, preferences, and expertise, as captured by their user histories $\mathbb{H}_\mathbb{U}$ and roles $\mathbb{C}$.

%By conditioning the language model's output on these additional factors, the PSW benchmark allows for the evaluation of personalized models that can adapt to the unique characteristics of individual authors in a collaborative writing environment.

\subsection{Task Description}

\paragraph{UP-0: Research Interest Generation:}
Before all the PSW tasks, we create a benchmark for user profiling. This involves compiling a list of research interests that accurately reflect each author's expertise and research focus based on their publication history. To acquire the necessary information, we extract the research interests of each author from Google Scholar\footnote{\url{https://github.com/scholarly-python-package/scholarly}} by searching their name. 
%Once we have this information, we incorporate it into the author's profile.
\vspace{-.25cm}

\paragraph{PSW-1: Research Topic Generation:}
This task aims to generate a list of research topics that capture the collaborating authors' joint expertise and research focus, given their user profiles. The generated research topics should be relevant to the authors' past publications and help identify potential research directions for their collaborative work. We use OpenAI's \textit{GPT-4} model to automatically extract research topics from selected papers. The extracted topics are then linked to their respective papers and author profiles.
\vspace{-.25cm}

\paragraph{PSW-2: Research Question Generation:}
This task focuses on generating a set of research questions that align with the expertise and interests of the collaborating authors and are relevant to the target paper. The generated research questions should help guide the content and structure of the collaborative writing process. We automatically use OpenAI's \textit{GPT-4} model to extract research questions from the selected papers for this task. The extracted research questions are then linked to their papers and author profiles.
\vspace{-.25cm}

\paragraph{PSW-3: Paper Abstract Generation:}
This task involves generating a paper abstract that summarizes the key points and contributions of the collaborative research paper, given the user profiles, research interests, target paper title, and research questions. 
%The generated abstract should incorporate the writing styles and preferences of the collaborating authors while maintaining coherence and clarity. For this task, w
We directly retrieve the abstracts from the selected papers using the Semantic Scholar API \footnote{\url{https://api.semanticscholar.org/}}. The retrieved abstracts are then linked to their respective papers and author profiles.
\vspace{-.25cm}

\paragraph{PSW-4: Paper Title Generation:}
This task aims to generate a suitable title for the collaborative research paper, considering the user profiles, research interests, research questions, and paper abstract. 
%The generated title should be concise, informative, and reflect the paper's main contributions while considering the collaborating authors' expertise and interests.
The data is collected by Semantic Scholar API as well.
\subsection{GPT-based Evaluation}
LLM-based evaluators, such as G-Eval \citep{liu2023gpteval}, have shown high consistency with human evaluators \cite{chang2024survey}, particularly in personalized text generation \cite{wang2023automated}. 
Therefore, we utilize GPT-4-turbo with chain-of-thought prompting as a judge to evaluate the generated outputs on the PSW benchmark in multiple dimensions \citep{zhang2019bertscore}, including consistency, fluency, relevance, and novelty. An example of our evaluation (G-Eval) prompt can be found in Appendix \ref{sec:prompt_geval}.

\vspace{-.1cm}

\section{Experimental Setup}

\vspace{-.1cm}

%We assess our methods alongside other baseline approaches across the LaMP and PSW datasets. This section provides a detailed exploration of the experimental settings for these evaluations.

\subsection{Datasets and Evaluation}

\textit{LaMP Dataset}
We follow the established LAMP benchmark \citet{salemi2023lamp}, encompassing three classification and four text generation tasks. 
Specifically, these tasks are Personalized Citation Identification (LaMP-1), Personalized News Categorization (LaMP-2), Personalized Product Rating (LaMP-3), Personalized News Headline Generation (LaMP-4), Personalized Scholarly Title Generation (LaMP-5), Personalized Email Subject Generation (LaMP-6), and Personalized Tweet Paraphrasing (LaMP-7).

\noindent
\textit{PSW Dataset}
%As introduced in the previous section, the PSW dataset is designed to assess the performance of personalized language models in collaborative scientific writing scenarios. 
The dataset includes one individual task, User Profiling (UP-0), and four collaborative tasks: Research Topics Generation (PSW-1), Research Question Generation (PSW-2), Paper Abstract Generation (PSW-3), and Paper Title Generation (PSW-4).

\vspace{-.1cm}

%\paragraph{Evaluation:}
Our evaluation follows the LaMP \cite{salemi2023lamp} and we employ the metrics specified in the LaMP for each task. Those include F1, Accuracy, MAE, and RMSE for classification tasks and ROUGE-1 and ROUGE-L for generation tasks.
%\subsection{Methods to compare}
%We employ \textit{GPT-3.5-turbo} for all tasks in this paper. 
We compare several baselines, including non-personalized language models, models fine-tuned on history data without personalization, and models that use a simple concatenation of user histories for personalization with retrieval models.

\vspace{-.1cm}

\subsection{Main Result}
\label{sec:main_result}

\begin{table}[t]
  \setlength{\tabcolsep}{1pt}
    \scriptsize
    \centering
  \scalebox{0.93}{
    \begin{tabular}{ccccccc}
      \toprule
      \rule{0pt}{1.2em}  &  & \multicolumn{2}{c}{\textbf{Non-personalized}} & \multicolumn{2}{c}{\textbf{Personalized$^\S$}} & \multirow{2}*{{Ours}}  
      \\
      \cmidrule(lr){3-4} \cmidrule(lr){5-6}
      \rule{0pt}{1.1em} \textbf{Dataset} & \textbf{Metric} & \textbf{FlanT5-XXL}$^\dag$ & \textbf{ChatGPT}$^\dag$ & \textbf{FlanT5-XXL}$^\dag$ & \textbf{ChatGPT}$^\dag$  \\
           \midrule \\[-1em]
      LaMP-1 & Accuracy & 0.522 & 0.510 & 0.675 & \textbf{0.701} & 0.624 \\
      
    \midrule  \\[-1em]
    \multirow{2}*{{LaMP-2}} & Accuracy & 0.591  & 0.610 & 0.598 & 0.693 & \textbf{0.729} \\
     & F1 & 0.463  & 0.455 & 0.477 & 0.455 & \textbf{0.591} \\
     
    \midrule  \\[-1em]
    \multirow{2}*{{LaMP-3}} & MAE & 0.357  & 0.699 & 0.282 & 0.658 & \textbf{0.274} \\
     & RMSE & 0.666  & 0.977 & 0.584 & 1.102 & \textbf{0.559} \\
     
    \midrule  \\[-1em]
    \multirow{2}*{{LaMP-4}} & ROUGE-1 & 0.164  & 0.133 & 0.192 & 0.160 & \textbf{0.195} \\
     & ROUGE-L & 0.149  & 0.118 & 0.178 & 0.142 & \textbf{0.180} \\
     
    \midrule  \\[-1em]
    \multirow{2}*{{LaMP-5}} & ROUGE-1 & 0.455  & 0.395 & 0.467 & 0.398 & \textbf{0.469 }\\
     & ROUGE-L & 0.410  & 0.334 & 0.424 & 0.336 & \textbf{0.426} \\

    \midrule  \\[-1em]
    \multirow{2}*{{LaMP-6}} & ROUGE-1 & 0.332  & - & 0.466 & - & \textbf{0.485 }\\
     & ROUGE-L & 0.320  & - & 0.453 & - & \textbf{0.464 }\\
     
    \midrule  \\[-1em]
    \multirow{2}*{{LaMP-7}} & ROUGE-1 & \textbf{0.459}  & 0.396 & 0.448 & 0.391 & 0.455 \\
     & ROUGE-L & \textbf{0.404}  & 0.337 & 0.396 & 0.324 & 0.398 \\
     
      \bottomrule
    \end{tabular}
    }
    \vspace{-.1cm}
    \caption{\textbf{Performance comparison of models on the LaMP dataset.} $^\dag$Baseline results are obtained directly from \cite{salemi2023lamp}. $^\S$ Personalized means we use retrieval modules before LLMs.}
    \label{tab:lamp}
    \vspace{-.6cm}
\end{table}

\textbf{LaMP Results} To ensure a fair comparison, we utilize a user-based separation from LaMP \cite{salemi2023lamp}. We only grant the model access to the provided user history and restrict it from accessing any other information. Additionally, we utilize \textit{the same pre-trained retriever in LaMP baselines}, without any additional fine-tuning, to retrieve the top five examples. This approach is identical to the \textit{Non-Personalized} setting in \cite{salemi2023lamp}. Finally, we compare our results with the outcomes reported in the study.

As shown in Table \ref{tab:lamp}, our analysis unveils a notable performance enhancement through our method's application, significantly when leveraging the same backbone language models (\textit{GPT-3.5-turbo}). 
It is clear that our ``gist''-style information compression is much more necessary than retrieval methods as the comparisons in Table \ref{tab:lamp}.
In the domain of text generation tasks (LaMP-4$\sim$7), our method achieves an average improvement of \textbf{$0.048$} in Rouge-1 and $0.053$ in Rouge-L, corresponding to gains of \textbf{$15.2\%$} and \textbf{$19.5\%$}, respectively. Similarly, for the classification tasks (LaMP-1$\sim$3), we observe an average \textbf{+$12.6\%$} accuracy gain of and a \textbf{+$42.5\%$} reduction in MAE compared to the \textit{Non-Personalized} setting. Our method continues to exhibit better performance across most tasks, even when compared with \textbf{\texttt{FlanT5-XXL}}, with a fine-tuned retriever as \textit{Personalized} setting. The prompt used in this experiment is detailed in Appendix \ref{sec:prompt_lamp}.

\begin{table}[!hbt]
\setlength\tabcolsep{1pt}
\scriptsize
\centering
\scalebox{0.8}{
\begin{tabular}{llcccccc}
\toprule
& & \multicolumn{6}{c}{\textbf{Metrics}} \\
\cmidrule(lr){3-8}
\textbf{Datasets} & \textbf{Method} & \textbf{ROUGE-1} & \textbf{ROUGE-L} & \textbf{Consistency} & \textbf{Fluency} & \textbf{Relevance} & \textbf{Novelty} \\
\midrule
UP-0 & \textbf{\texttt{Single-Author}} & 0.267 & 0.233 & 4.32 & 2.01 & 3.59 & / \\
\midrule
\multirow{3}{*}{PSW-1} & \textbf{\texttt{Zero-shot}} & 0.306 & 0.257 & 3.43 & \textbf{2.65} & 3.53 & 2.30 \\
& \textbf{\texttt{Single-Author}} & 0.325 & 0.266 & 3.44 & 2.47 & 3.61 & 2.59 \\
& \textbf{\texttt{Multi-Author}} & \textbf{0.337} & \textbf{0.280} & \textbf{3.59} & 2.58 & \textbf{3.67} & \textbf{2.63} \\
\midrule
\multirow{3}{*}{PSW-2} & \textbf{\texttt{Zero-shot}} & 0.196 & 0.179 & 4.31 & 2.04 & 3.89 & 2.21 \\
& \textbf{\texttt{Single-Author}} & 0.190 & 0.171 & 4.20 & 2.23 & 3.67 & 2.01 \\
& \textbf{\texttt{Multi-Author}} & \textbf{0.201} & \textbf{0.186} & \textbf{4.60} & \textbf{2.39} & \textbf{3.91} & \textbf{2.38} \\
\midrule
\multirow{3}{*}{PSW-3} & \textbf{\texttt{Zero-shot}} & 0.099 & 0.094 & 4.43 & 2.81 & 4.43 & 2.40 \\
& \textbf{\texttt{Single-Author}} & 0.131 & 0.124 & \textbf{4.94} & \textbf{2.94} & 4.70 & 2.40 \\
& \textbf{\texttt{Multi-Author}} & \textbf{0.145} & \textbf{0.131} & 4.92 & \textbf{2.94} & \textbf{4.71} & \textbf{2.45} \\
\midrule
\multirow{3}{*}{PSW-4} & \textbf{\texttt{Zero-shot}} & 0.459 & 0.391 & 4.41 & 2.41 & 3.58 & 2.38 \\
& \textbf{\texttt{Single-Author}} & 0.472 & 0.409 & 4.59 & 2.49 & 3.78 & 2.60 \\
& \textbf{\texttt{Multi-Author}} & \textbf{0.505} & \textbf{0.444} & \textbf{4.64} & \textbf{2.59} & \textbf{3.79} & \textbf{2.64} \\
\bottomrule
\end{tabular}
}
    \vspace{-.1cm}
\caption{\textbf{Performance comparison of personalized models on the PSW dataset}. We report additional metrics such as \textbf{Consistency (1-5)}, \textbf{Fluency (1-3)}, \textbf{Relevance (1-5)}, and \textbf{Novelty (1-3)}.}
\label{tab:psw_results}
\vspace{-.5cm}
\end{table}

\textbf{PSW Results:}
We then evaluate our proposed model using the PSW dataset, focusing on user profiling (UP-0), personalized idea brainstorming (PSW-1, PSW-2), and personalized text generation (PSW-3, PSW-4) in three different settings:
\begin{enumerate}[topsep=0pt,parsep=1pt]

\item \textbf{\texttt{Zero-shot}}: Generates outputs based on the input prompt $x$ alone:
$
y = \operatorname{Generate}(x)$.

\vspace{-.2cm}

\item \textbf{\texttt{Single-Author}}: Personalizes with single user's profile $P_{u_i}$ and retrieved snippets $R_i$:
$
y = \operatorname{Generate}(\tilde{x}, P_{u_i})$,
where $\tilde{x} = [x; \mathbb{R}_i]$ and $\mathbb{R}_i = \operatorname{Retrieve}(x, \mathbb{H}_i, 10)$.

\vspace{-.2cm}

\item \textbf{\texttt{Multi-Author}}: Personalizes with multiple users' profiles $\mathbb{P}_\mathbb{U}$ and retrieved snippets $\mathbb{R}$:
$
y = \operatorname{Generate}(\tilde{x}, \mathbb{P}_\mathbb{U})$,
where $\tilde{x} = [x; \mathbb{R}_1; \cdots; \mathbb{R}_n]$, $\mathbb{R}_i = \operatorname{Retrieve}(x, \mathbb{H}_i, 10)$ for each user $u_i$.

\end{enumerate}

As shown in Table \ref{tab:psw_results}, our \textbf{\texttt{Multi-Author}} setting demonstrates superior performance across all tasks. In PSW-1 and PSW-2, the \textbf{\texttt{Multi-Author}} setting outperforms both \textbf{\texttt{Zero-shot}} and \textbf{\texttt{Single-Author}} settings, with an average improvement of \textbf{+$6.9\%$} in ROUGE-1 and \textbf{+$7.1\%$} in ROUGE-L. 
Similarly, for the PSW-3 and PSW-4, the \textbf{\texttt{Multi-Author}} setting achieves the highest ROUGE scores, with an average gain of \textbf{+$28.2\%$} in ROUGE-1 and \textbf{+$26.6\%$} in ROUGE-L, compared to the \textbf{\texttt{Zero-shot}} and \textbf{\texttt{Single-Author}} settings. 
Furthermore, the \textbf{\texttt{Multi-Author}} setting exhibits the highest scores for additional metrics such as Consistency, Fluency, Relevance, and Novelty across all tasks, with an average improvement of \textbf{+$5.1\%$}, \textbf{+$6.7\%$}, \textbf{+$3.8\%$}, and \textbf{+$6.4\%$}, respectively, compared to the \textbf{\texttt{Zero-shot}} and \textbf{\texttt{Single-Author}} setting. The prompt used in this experiment is detailed in Appendix \ref{sec:prompt_psw}.

%Table \ref{tab:ablation_order} and \ref{tab:ablation_profile}  report the result

\subsection{Ablation Study}
Finally, to evaluate the contribution of each component, we perform an ablation study when: 1) Switching the order of users and 2) Removing user profiling.

\subsubsection{Impact of Author Order}

\label{app:ablation_order}
Table \ref{tab:ablation_order} shows how changing the author order affects the performance of multi-user personalized models. We experiment with three variants:

 \textbf{\texttt{Original}}: The original author order as provided in the dataset. \textbf{\texttt{Swap-Random}}: Randomly shuffle the order of authors.
\textbf{\texttt{Swap-First}}: Move the first author to the end of the author list.

\begin{table}[!hbt]
\setlength{\tabcolsep}{1pt}
\scriptsize
\centering
\scalebox{0.8}{
\begin{tabular}{llcccccc}
\toprule
& & \multicolumn{6}{c}{\textbf{Metrics}} \\
\cmidrule(lr){3-8}
\textbf{Datasets} & \textbf{Variants} & \textbf{ROUGE-1} & \textbf{ROUGE-L} & \textbf{Consistency} & \textbf{Fluency} & \textbf{Relevance} & \textbf{Novelty} \\
\midrule
\multirow{3}{*}{PSW-1} & \textbf{\texttt{Original}} & \textbf{0.337} & \textbf{0.280} & \textbf{3.59} & \textbf{2.58} & 3.67 & \textbf{2.63} \\
& \textbf{\texttt{Swap-Random}} & 0.321 & 0.272 & 3.42 & 2.48 & \textbf{3.69} & 2.45 \\
& \textbf{\texttt{Swap-First}} & 0.314 & 0.260 & 3.35 & 2.42 & 3.48 & 2.37 \\
\midrule
\multirow{3}{*}{PSW-2} & \textbf{\texttt{Original}} & \textbf{0.201} & \textbf{0.186} & \textbf{4.60} & \textbf{2.39} & \textbf{3.91} & 2.38 \\
& \textbf{\texttt{Swap-Random}} & 0.193 & 0.178 & 4.53 & 2.30 & 3.85 & \textbf{2.42} \\
& \textbf{\texttt{Swap-First}} & 0.186 & 0.171 & 4.46 & 2.27 & 3.77 & 2.29 \\
\midrule
\multirow{3}{*}{PSW-3} & \textbf{\texttt{Original}} & \textbf{0.145} & \textbf{0.131} & \textbf{4.92} & 2.94 & \textbf{4.71} & \textbf{2.45} \\
& \textbf{\texttt{Swap-Random}} & 0.138 & 0.125 & 4.84 & 2.88 & 4.65 & 2.50 \\
& \textbf{\texttt{Swap-First}} & 0.130 & 0.117 & 4.78 &\textbf{2.98} & 4.57 & 2.55 \\
\midrule
\multirow{3}{*}{PSW-4} & \textbf{\texttt{Original}} & \textbf{0.505} & \textbf{0.444} & \textbf{4.64} & \textbf{2.59} & \textbf{3.79} & 2.64 \\
& \textbf{\texttt{Swap-Random}} & 0.492 & 0.431 & 4.57 & 2.55 & 3.72 & 2.70 \\
& \textbf{\texttt{Swap-First}} & 0.483 & 0.421 & 4.50 & 2.50 & 3.64 & \textbf{2.76} \\
\bottomrule
\end{tabular}
}
\caption{\textbf{Impact of author order on the performance of multi-user personalized models} We report additional metrics such as \textbf{Consistency (1-5)}, \textbf{Fluency (1-3)}, \textbf{Relevance (1-5)}, and \textbf{Novelty (1-3)}.}
\label{tab:ablation_order}
\vspace{-.5cm}
\end{table}

The \textbf{\texttt{Original}} order consistently achieves the best performance across all metrics on all PSW tasks. Randomly swapping authors (\textbf{\texttt{Swap-Random}}) leads to a slight decline, while moving the first author to the end (\textbf{\texttt{Swap-First}}) results in a more significant drop. This observation highlights the importance of preserving the original author order in multi-author collaborative writing scenarios. The first author, often the lead or corresponding author, significantly influences the document's content, structure, and style. As a result, their writing style and expertise tend to be most prominently reflected in the document. Disrupting this order introduces noise and hinders the model's ability to capture the individual authors' impact and the logical progression of ideas, particularly affecting the generation tasks (PSW-3 and PSW-4), where content and style are heavily influenced by the main author's expertise and preferences.

\subsubsection{Impact of User Profiling}
\label{app:ablation_profile}

Table \ref{tab:ablation_profile} reports ablation results on the user profile component:

\textbf{\texttt{Original}}: User profiles constructed using \OURS{}.
\textbf{\texttt{Removed}}: No user profiles were used, only retrieving relevant snippets.
\textbf{\texttt{Random}}: Replacing target user profiles with randomly sampled user profiles.

\begin{table}[!hbt]
\setlength{\tabcolsep}{1pt}
\scriptsize
\centering
\scalebox{0.8}{
\begin{tabular}{llcccccc}
\toprule
& & \multicolumn{6}{c}{\textbf{Metrics}} \\
\cmidrule(lr){3-8}
\textbf{Datasets} & \textbf{Profile} & \textbf{ROUGE-1} & \textbf{ROUGE-L} & \textbf{Consistency} & \textbf{Fluency} & \textbf{Relevance} & \textbf{Novelty} \\
\midrule
\multirow{3}{*}{PSW-1} & \textbf{\texttt{Original}}    & \textbf{0.337} & \textbf{0.280} & \textbf{3.59} & \textbf{2.58} & \textbf{3.67} & \textbf{2.63} \\
& \textbf{\texttt{Removed}}     & 0.297 & 0.250 & 3.21 & 2.49 & 3.31 & 2.57 \\
& \textbf{\texttt{Random}}      & 0.328 & 0.272 & 3.55 & 2.56 & 3.62 & 2.68 \\
\midrule
\multirow{3}{*}{PSW-2} & \textbf{\texttt{Original}}    & \textbf{0.201} & \textbf{0.186} & \textbf{4.60} & 2.39 & \textbf{3.91} & 2.38 \\
& \textbf{\texttt{Removed}}     & 0.180 & 0.166 & 4.28 & 2.32 & 3.63 & 2.33 \\
& \textbf{\texttt{Random}}      & 0.195 & 0.182 & 4.57 & \textbf{2.42} & 3.89 & \textbf{2.45} \\
\midrule

\multirow{3}{*}{PSW-3} & \textbf{\texttt{Original}}    & \textbf{0.145} & \textbf{0.131} & 4.92 & 2.94 & \textbf{4.71} & 2.45 \\
& \textbf{\texttt{Removed}}     & 0.128 & 0.115 & 4.70 & 2.87 & 4.50 & 2.41 \\
& \textbf{\texttt{Random}}      & 0.142 & 0.128 & \textbf{4.95} & \textbf{2.96} & 4.69 & \textbf{2.51} \\
\midrule
\multirow{3}{*}{PSW-4} & \textbf{\texttt{Original}}    & \textbf{0.505} & \textbf{0.444} & \textbf{4.64} & \textbf{2.59} & \textbf{3.79} & 2.64 \\
& \textbf{\texttt{Removed}}     & 0.475 & 0.419 & 4.38 & 2.53 & 3.58 & 2.56 \\
& \textbf{\texttt{Random}}      & 0.498 & 0.438 & 4.60 & 2.58 & 3.76 & \textbf{2.69} \\

\bottomrule
\end{tabular}}
\caption{\textbf{Impact of the user profile on the performance of multi-user personalized models.} We report additional metrics such as \textbf{Consistency (1-5)}, \textbf{Fluency (1-3)}, \textbf{Relevance (1-5)}, and \textbf{Novelty (1-3)}.}
\label{tab:ablation_profile}
\vspace{-.5cm}
\end{table}

Removing user profiles (\textbf{\texttt{Removed}}) leads to the largest performance decline, confirming the benefit of \OURS{} in multi-user personalization. Using random profile texts (\textbf{\texttt{Random}}) recovers some of the gaps but still underperforms the \textbf{\texttt{Original}} profiles. This demonstrates that the distilled user traits successfully capture useful information for collaborative writing, such as individual writing styles, expertise, and preferences. The performance gap between \textbf{\texttt{Original}} and \textbf{\texttt{Random}} profiles highlights the effectiveness of the \OURS{} technique in extracting relevant user characteristics from their background information. These findings underscore the importance of incorporating author-specific traits to enable a more personalized and contextually appropriate generation in multi-user settings.

\section{Conclusion}
In summary, we introduce a training-free technique, \OURS{}, for personalizing large language models by distilling user interactions using gist into concise profiles.
Moreover, we extend the LaMP dataset into the Personalized Scientific Writing (PSW) dataset to evaluate multi-user scenarios in collaborative scientific writing.
Our experiments show that the proposed method is effective on the LaMP and PSW datasets. In particular, both single-user and multi-user settings validate the benefits of profile-guided personalization.
Finally, studying the interpretability and controllability of profile-guided models can help build user trust and allow for more fine-grained customization.

\section*{Limitation}
Our proposed \OURS{} framework has a few limitations that warrant discussion and could be addressed in future work:

\noindent
\textit{Dataset Specificity}
The experiments and results presented are primarily based on the Personalized Scientific Writing (PSW) dataset and the LaMP benchmark. While these datasets provide a diverse set of tasks, the performance and applicability of the \OURS{} framework may vary with different datasets or domains not covered by our experiments. Future work should evaluate the model on more varied datasets to ensure generalizability.

\noindent
\textit{Complexity of Profiles}
The profile generation process involves distilling user histories into concise representations. While this method captures essential traits, it may oversimplify user preferences and neglect nuanced behaviors present in longer and more complex histories. More sophisticated profiling techniques that can retain and effectively compress these complexities are needed.

% \noindent
% \textit{Scalability and Efficiency}
% Although the \OURS{} method improves memory management, the approach still has scalability concerns, particularly with very large user histories or an increasing number of collaborators. Efficiently managing and retrieving relevant user data from extensive histories without compromising performance remains a challenge.

\noindent
\textit{Scalability in Multi-Author Scenarios}
\july{The current implementation of \OURS{} may face challenges in scenarios with a very large number of authors, as individual personalities might become diluted or ``washed out" in the collective profile. This limitation could affect the effectiveness of personalization in large-scale collaborative projects. Future research should investigate the scalability of \OURS{} as the number of authors increases. By systematically varying the number of collaborators, we can determine how personalization effectiveness changes and at what point, if any, the benefits of individualized profiles diminish. This will help establish guidelines for applying \OURS{} across various collaborative scenarios.}

\noindent
\textit{Optimal Collaborative Writing Strategies}
\july{Our current study does not explore the optimal strategies for collaborative scientific writing using \OURS{}. To address this limitation, we propose future research aimed at identifying the most effective approaches to team-based scientific writing. This could involve comparing various team compositions, such as groups of diverse experts versus more homogeneous teams, or analyzing different collaborative processes enabled by \OURS{}. The research would aim to quantify how different collaborative strategies impact the quality, innovation, and efficiency of scientific writing. Findings from this study could provide valuable insights into best practices for team formation and collaboration in scientific writing, potentially leading to more effective use of AI-assisted tools like \OURS{} in academic and research contexts.}

\noindent
\textit{Dynamic Adaptation}
The current method creates static profiles based on available user histories at a given time. However, user preferences and styles may evolve, especially in dynamic collaborative environments. Developing a mechanism to update profiles dynamically based on real-time user interactions and feedback could further enhance the personalization capabilities.

\noindent
\textit{Evaluation Metrics}
The evaluation relies heavily on established metrics such as ROUGE and human-aligned scoring via G-Eval, which, while comprehensive, may not capture all dimensions of personalized content quality. Developing and employing more specialized evaluation metrics for personalized content generation, particularly in scientific and collaborative writing, would provide deeper insights into the effectiveness of the methods.

\noindent
\textit{Human Factors}:
Although tools like GPT-4 mitigate the involvement of human evaluation, it is inherently subjective. Future work should consider more robust and unbiased methods of human evaluation to validate the effectiveness of personalized outputs objectively.

\noindent
\textit{Ethical and Privacy Concerns}
Personalizing models using user histories raises potential ethical and privacy issues. It is crucial to ensure that user data is handled securely and that privacy concerns are adequately addressed. Future research should explore more privacy-preserving techniques for personalization, such as federated learning.

Adapting \OURS{} to long histories spanning multiple sessions is another valuable direction. 
Future work can explore more advanced profiling strategies, such as hierarchical representations and dynamic profile updates based on user feedback.

\section*{Ethical Statement}

\noindent
\textit{Dataset Licensing}
We have constructed the Personalized Scientific Writing (PSW) dataset, which will be publicly released under the MIT license. This permissive license allows users to freely use, modify, and distribute the dataset. By releasing the PSW dataset under the MIT license, we aim to promote transparency, reproducibility, and wide adoption of our research within the community.

\noindent
\textit{Artifact Use Consistent With Intended Use}
Regarding our use of existing artifacts, we have ensured that our usage is consistent with their intended purposes, as specified by their creators. For the artifacts we create, including the PSW dataset, we specify that the intended use is for research purposes. This is compatible with the original access conditions of any derivative data we utilized. Derivative data accessed for research purposes should not be used outside of research contexts.

\noindent
\textit{Personally-Identifying Info}
We acknowledge that the PSW dataset construction involved the use of researchers' real names to accurately reflect their contributions and expertise. However, to protect individual privacy and prevent any potential personal information leakage, the publicly released version of our dataset replaces real names with unique identifiers (IDs). This anonymization step ensures that no personally identifying information is disclosed while maintaining the dataset's utility for research purposes.

We have taken these steps to safeguard the privacy and personal information of the individuals whose data contributed to our research. Additionally, we have reviewed the dataset to ensure it does not contain any offensive content.

\noindent
\textit{Documentation Of Artifacts}
While our dataset does not involve artificial distributions, we have collected and included gender information in the metadata. This metadata, along with other relevant descriptive information about the dataset, will be made publicly available upon the paper's acceptance.

\bibliography{custom}

\appendix
\label{appendix}
\clearpage

\section{The PSW Dataset}
\label{sec:psw_overview}

\textbf{Overview.} The PSW dataset is constructed using data from the Semantic Scholar database \citep{fricke2018semantic}. We first selected a subset of papers from Software Engineering published after 2000, considering only papers with at least two authors to ensure the feasibility of evaluating collaborative writing scenarios. The collected papers were randomly split into training, validation, and test subsets.\footnote{We only used the test split in this paper since our method doesn't require model training.} We performed the split at the paper level to ensure that all tasks within the PSW benchmark had consistent data splits. The summary of PSW dataset statistics can be found in Table \ref{tab:psw_stats}.

\begin{table}[hbpt!]
\setlength{\tabcolsep}{4pt}

\centering
\scalebox{0.8}{
\begin{tabular}{lrrr}
\toprule
\textbf{Statistic} & \textbf{Train} & \textbf{Valid} & \textbf{Test} \\
\midrule
\# of Papers & 1,744 & 500 & 500 \\
\# of Authors & 6,461 & 1,655 & 1,280 \\
Avg. Authors / Paper & 4.05 & 3.16 & 3.25 \\
Avg. History Papers / Author & 63.47 & 75.34 & 92.21 \\
Avg. Research Interests / Author & 2.84 & 2.77 & 2.79 \\
\midrule
Avg. Title Length & 97.03 & 95.54 & 96.16 \\
Avg. Abstract Length & 970.92 & 981.36 & 1,037.09 \\
Avg. Research Question Length & 470.57 & 398.22 & 442.31 \\
\midrule
Avg. References / Paper & 60.24 & 54.85 & 58.93 \\
\bottomrule
\end{tabular}
}
\caption{\textbf{PSW Dataset Statistics with Train / Valid / Test Splits.}}
\label{tab:psw_stats}
\end{table}

\clearpage
\section{Metrics Visualization on PSW Dataset}
\begin{figure}[!hbt]
  \centering
  \begin{subfigure}[b]{0.45\linewidth}
    \includegraphics[width=\linewidth]{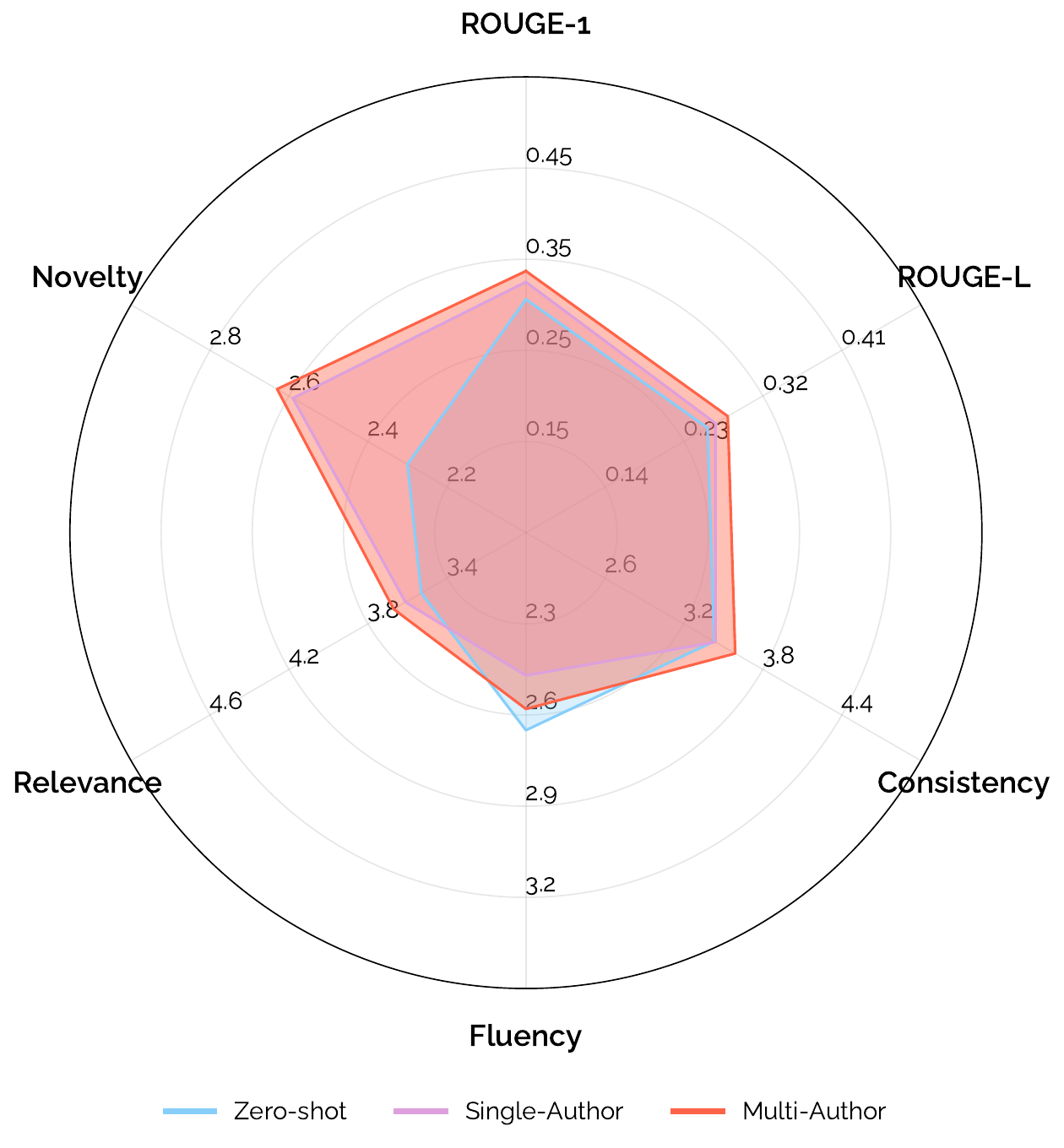}
    \caption{PSW-1}
    \label{fig:psw1}
  \end{subfigure}
  \hfill
  \begin{subfigure}[b]{0.45\linewidth}
    \includegraphics[width=\linewidth]{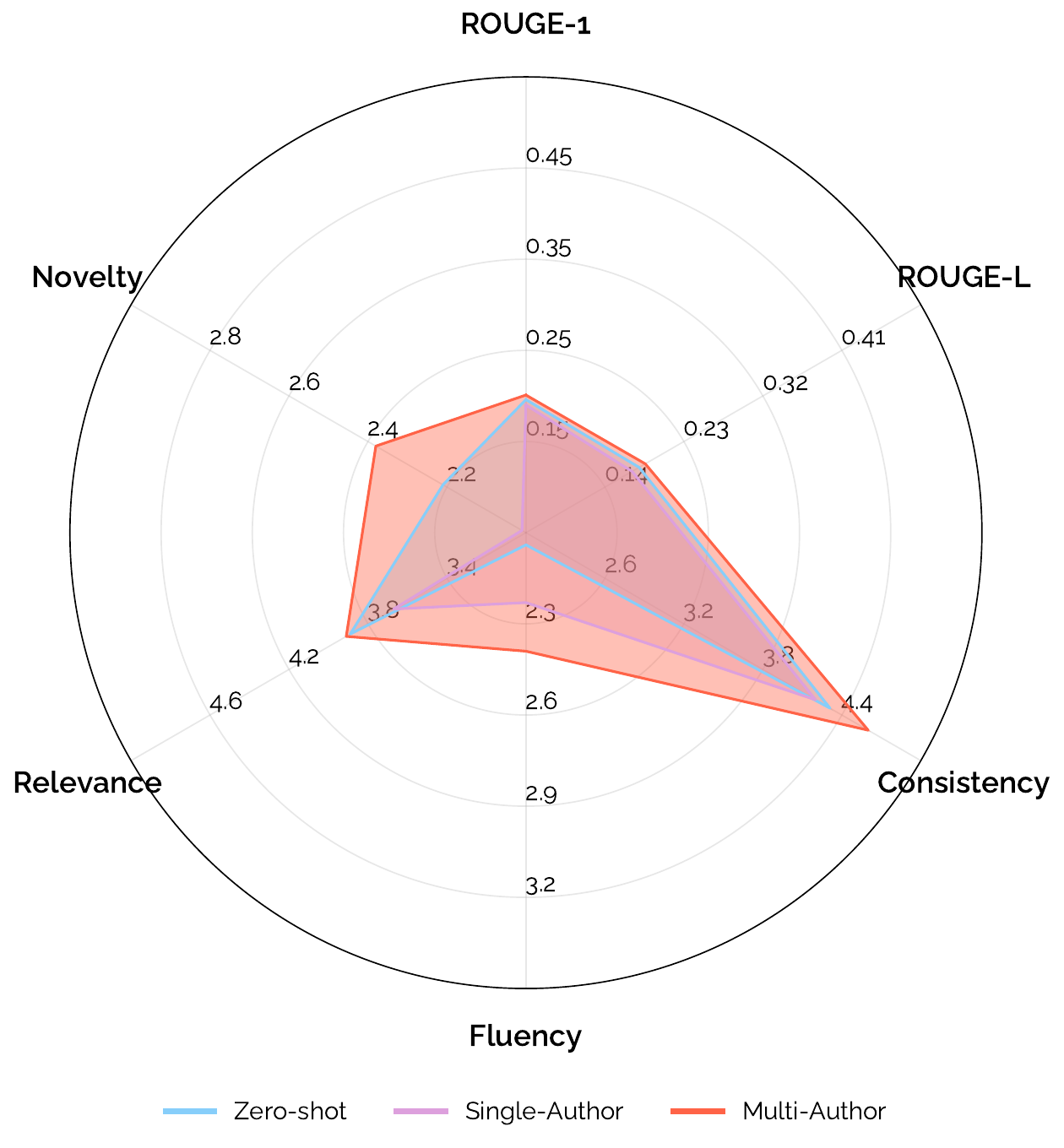}
    \caption{PSW-2}
    \label{fig:psw2}
  \end{subfigure}
  \hfill
  \begin{subfigure}[b]{0.45\linewidth}
    \includegraphics[width=\linewidth]{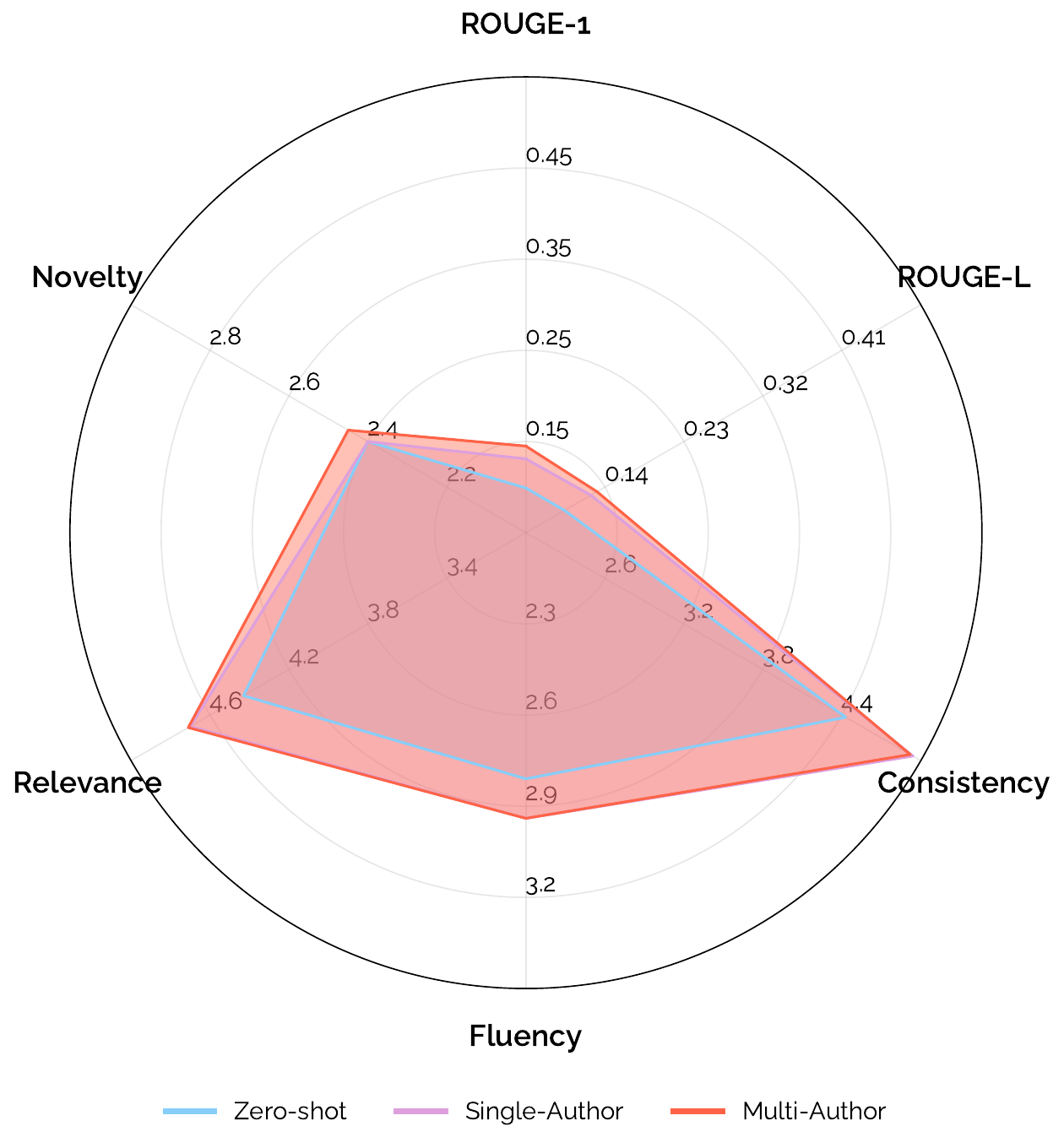}
    \caption{PSW-3}
    \label{fig:psw3}
  \end{subfigure}
  \hfill
  \begin{subfigure}[b]{0.45\linewidth}
    \includegraphics[width=\linewidth]{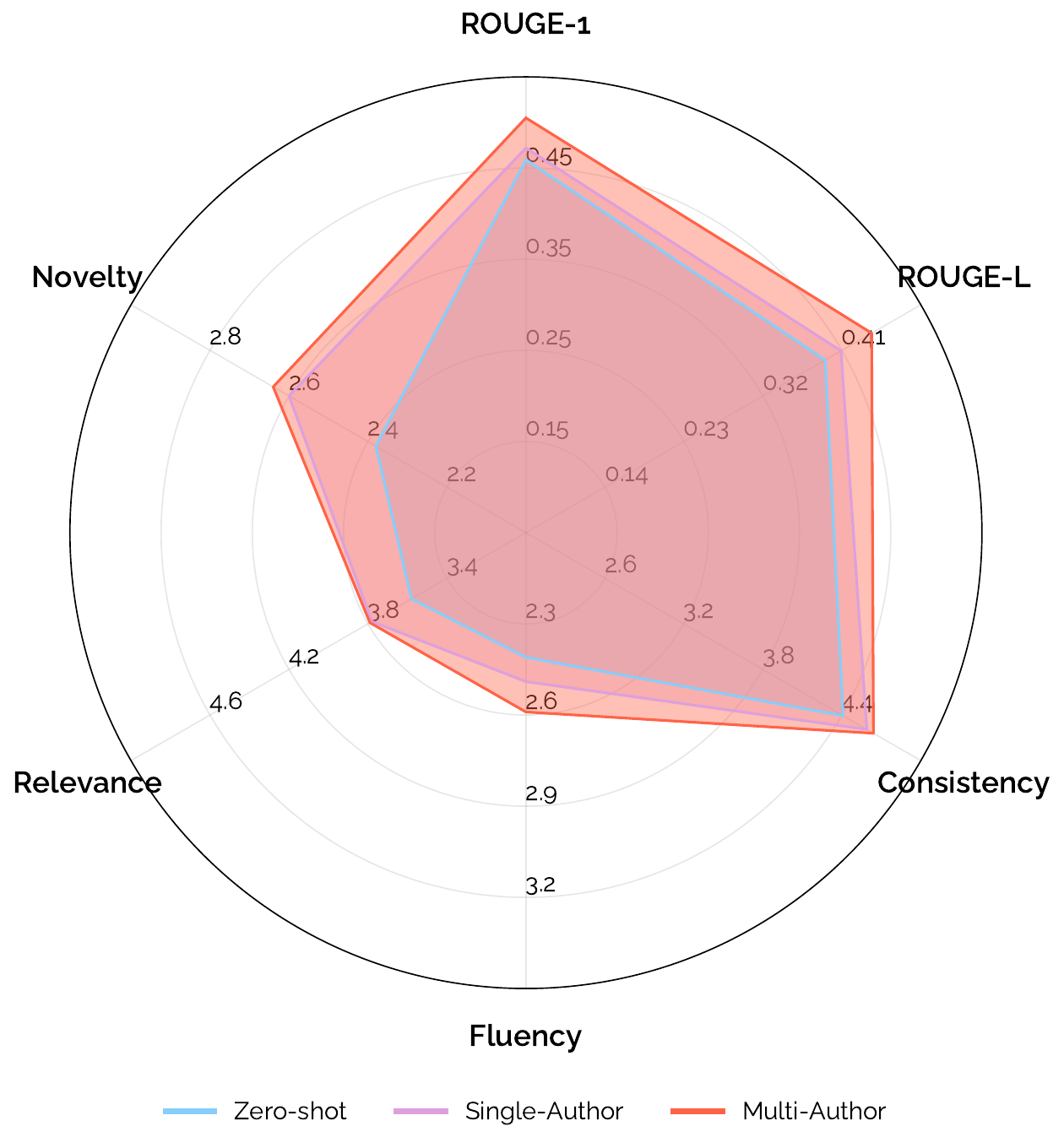}
    \caption{PSW-4}
    \label{fig:psw4}
  \end{subfigure}
  
  \caption{\textbf{Performance metrics across three different models}: \textbf{\texttt{Zero-shot}}, \textbf{\texttt{Single-Author}}, and \textbf{\texttt{Multi-Author}}. The \textbf{\texttt{Multi-Author}} model consistently achieves the highest scores across all datasets.}
  \label{fig:psw_main}
\end{figure}

\begin{figure}[!hbt]
  \centering
  \begin{subfigure}[b]{0.45\linewidth}
    \includegraphics[width=\linewidth]{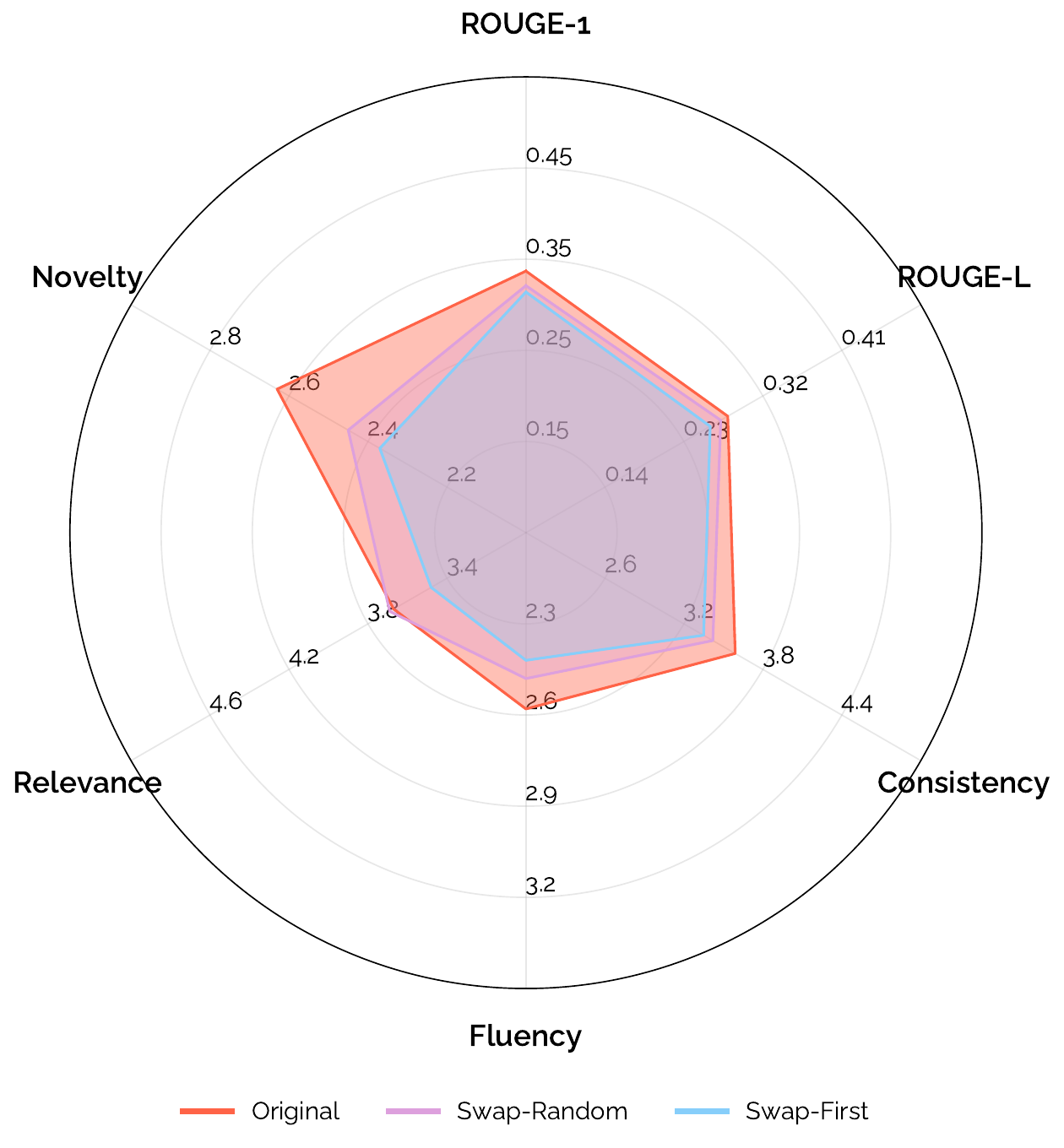}
    \caption{PSW-1}
    \label{fig:psw1-ablation1}
  \end{subfigure}
  \hfill
  \begin{subfigure}[b]{0.45\linewidth}
    \includegraphics[width=\linewidth]{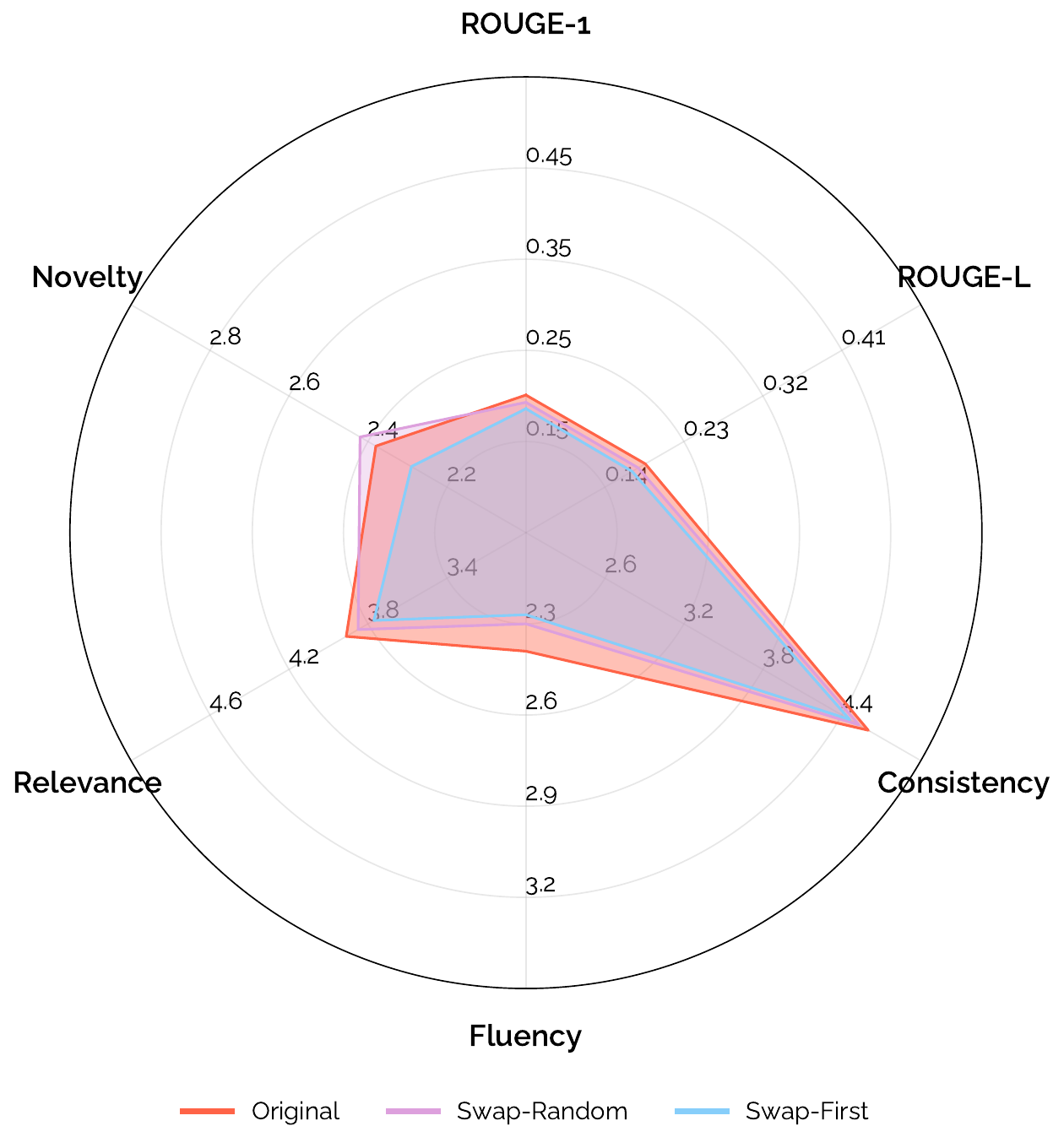}
    \caption{PSW-2}
    \label{fig:psw2-ablation1}
  \end{subfigure}
    \hfill
  \begin{subfigure}[b]{0.45\linewidth}
    \includegraphics[width=\linewidth]{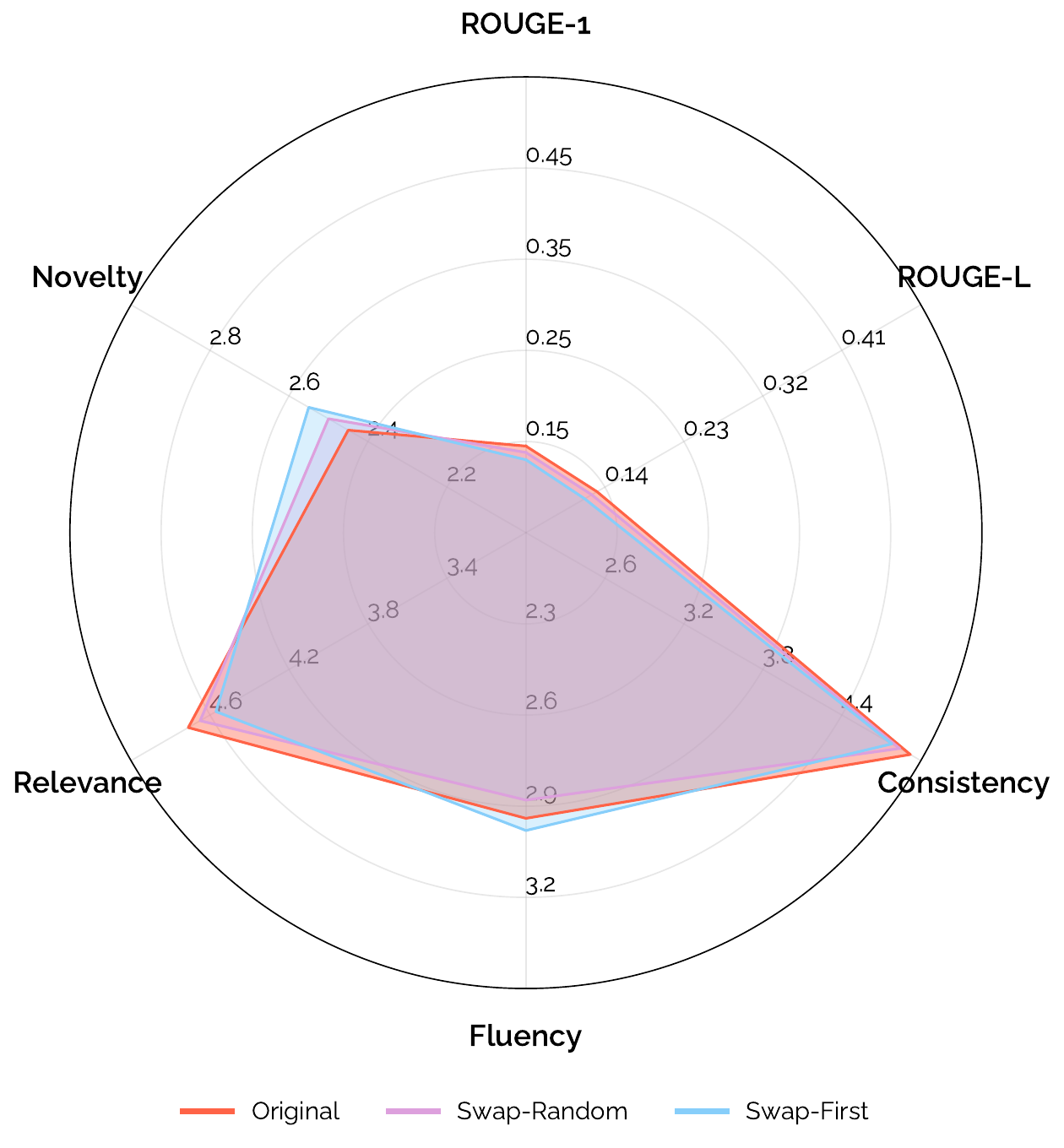}
    \caption{PSW-3}
    \label{fig:psw3-ablation1}
  \end{subfigure}
  \hfill
  \begin{subfigure}[b]{0.45\linewidth}
    \includegraphics[width=\linewidth]{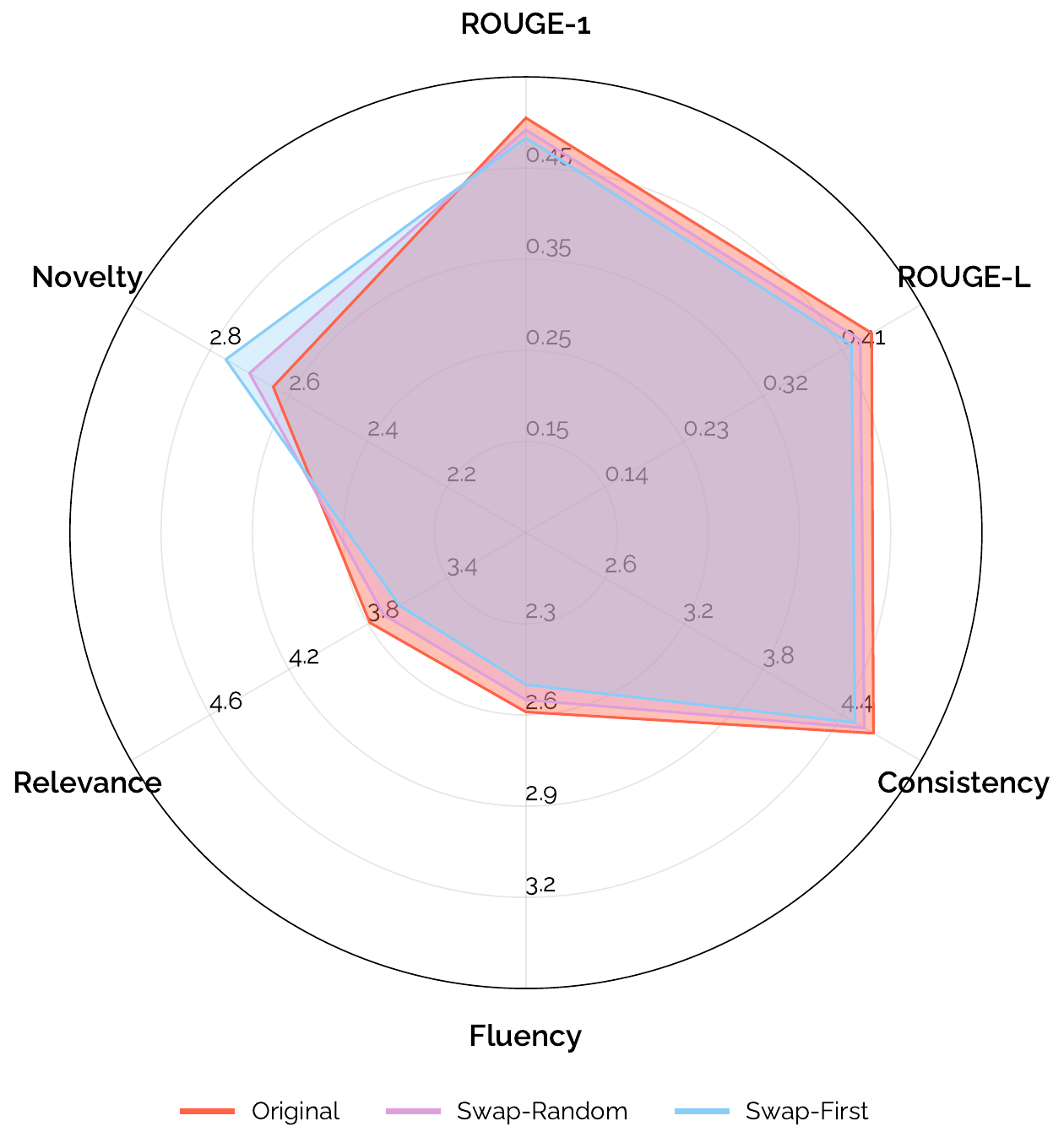}
    \caption{PSW-4}
    \label{fig:psw4-ablation1}
  \end{subfigure}
  
  \caption{\textbf{Impact of author order on the performance across three different models}: \textbf{\texttt{Original}}, \textbf{\texttt{Swap-Random}}, and \textbf{\texttt{Swap-First}}. The \textbf{\texttt{Original}} model consistently achieves the highest scores across all datasets.}
  \label{fig:psw_main-ablation1}
\end{figure}

\begin{figure}[!hbt]
  \centering
  \begin{subfigure}[b]{0.45\linewidth}
    \includegraphics[width=\linewidth]{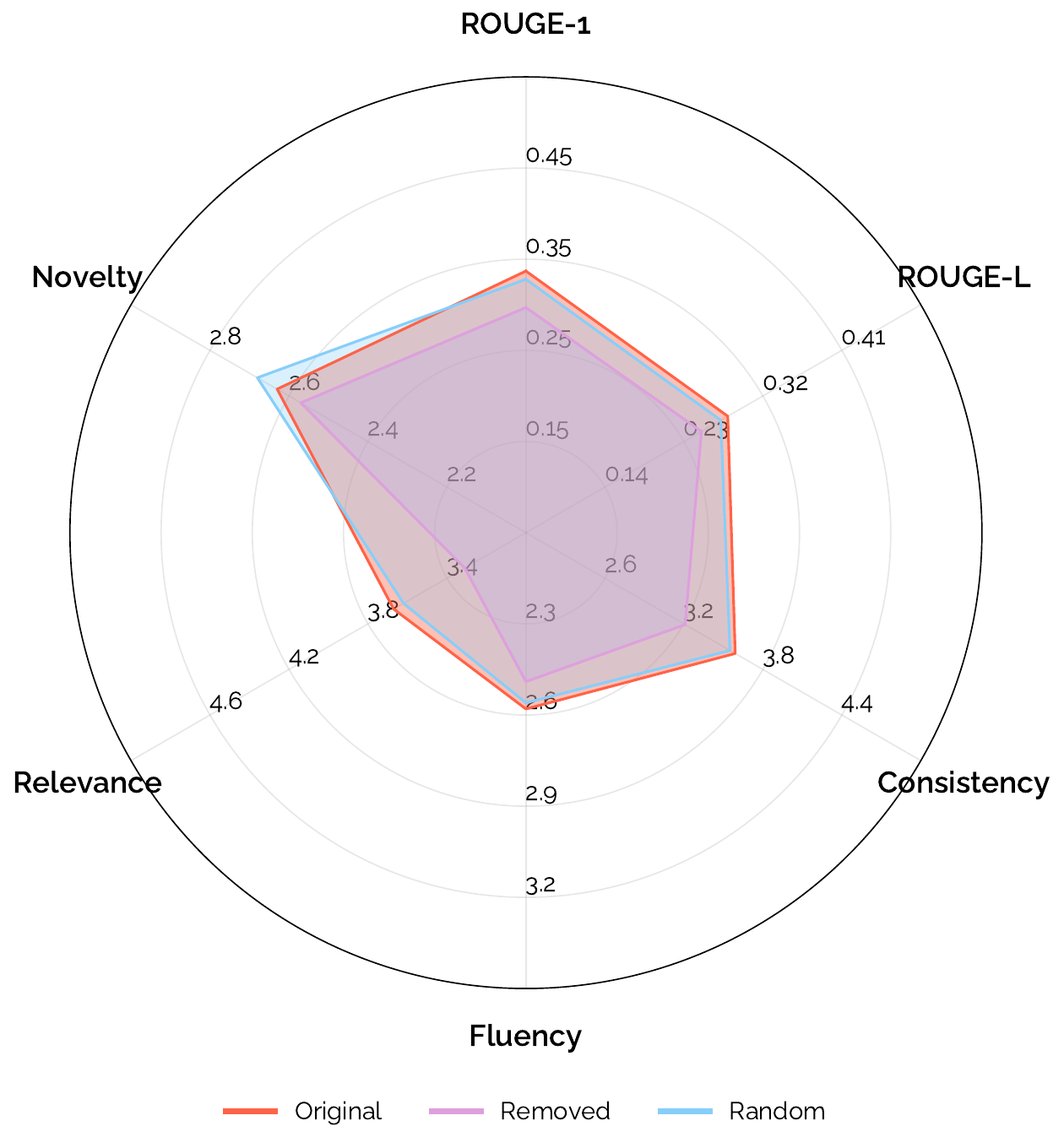}
    \caption{PSW-1}
    \label{fig:psw1-ablation2}
  \end{subfigure}
  \hfill
  \begin{subfigure}[b]{0.45\linewidth}
    \includegraphics[width=\linewidth]{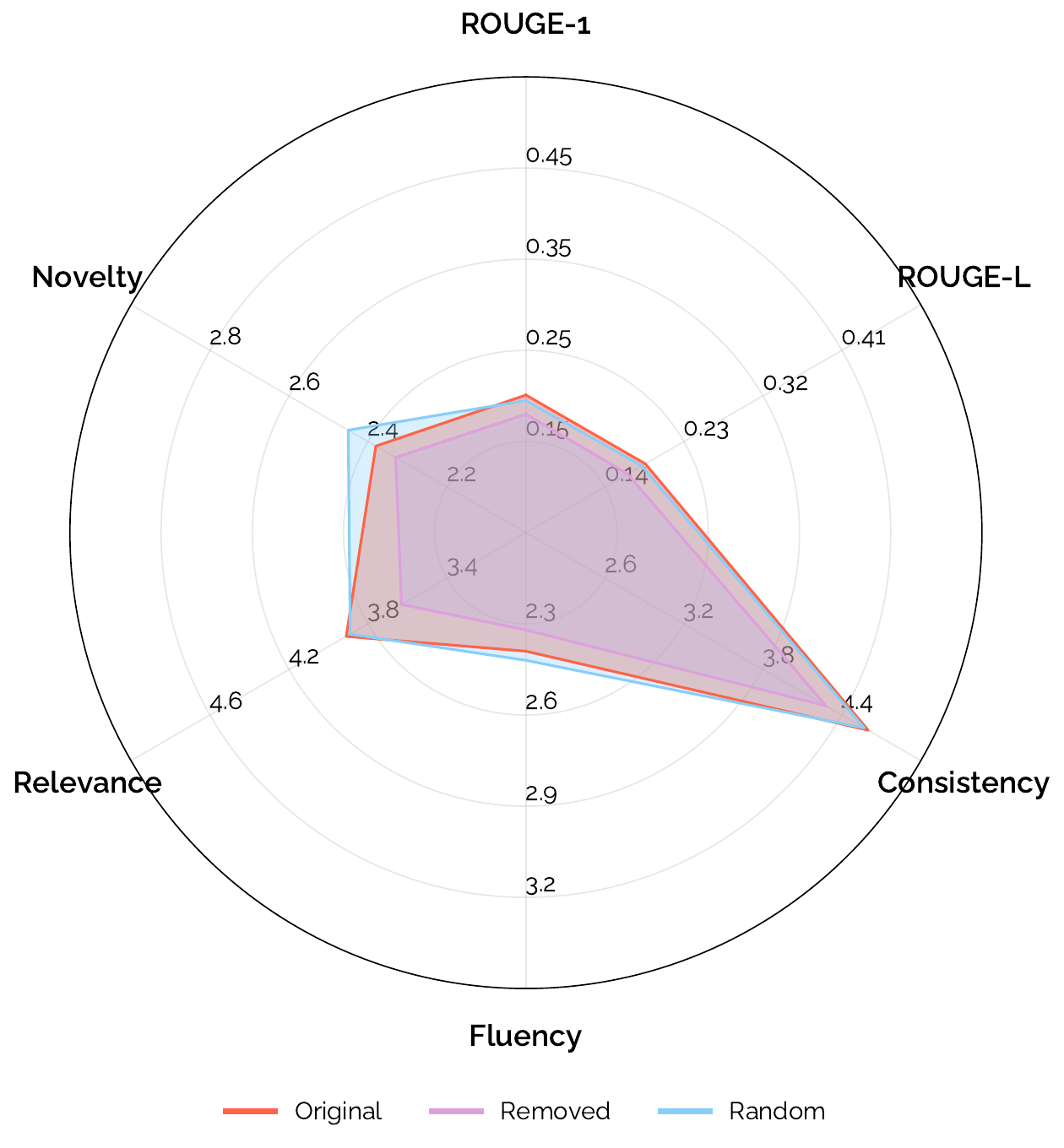}
    \caption{PSW-2}
    \label{fig:psw2-ablation2}
  \end{subfigure}
\hfill
  \begin{subfigure}[b]{0.45\linewidth}
    \includegraphics[width=\linewidth]{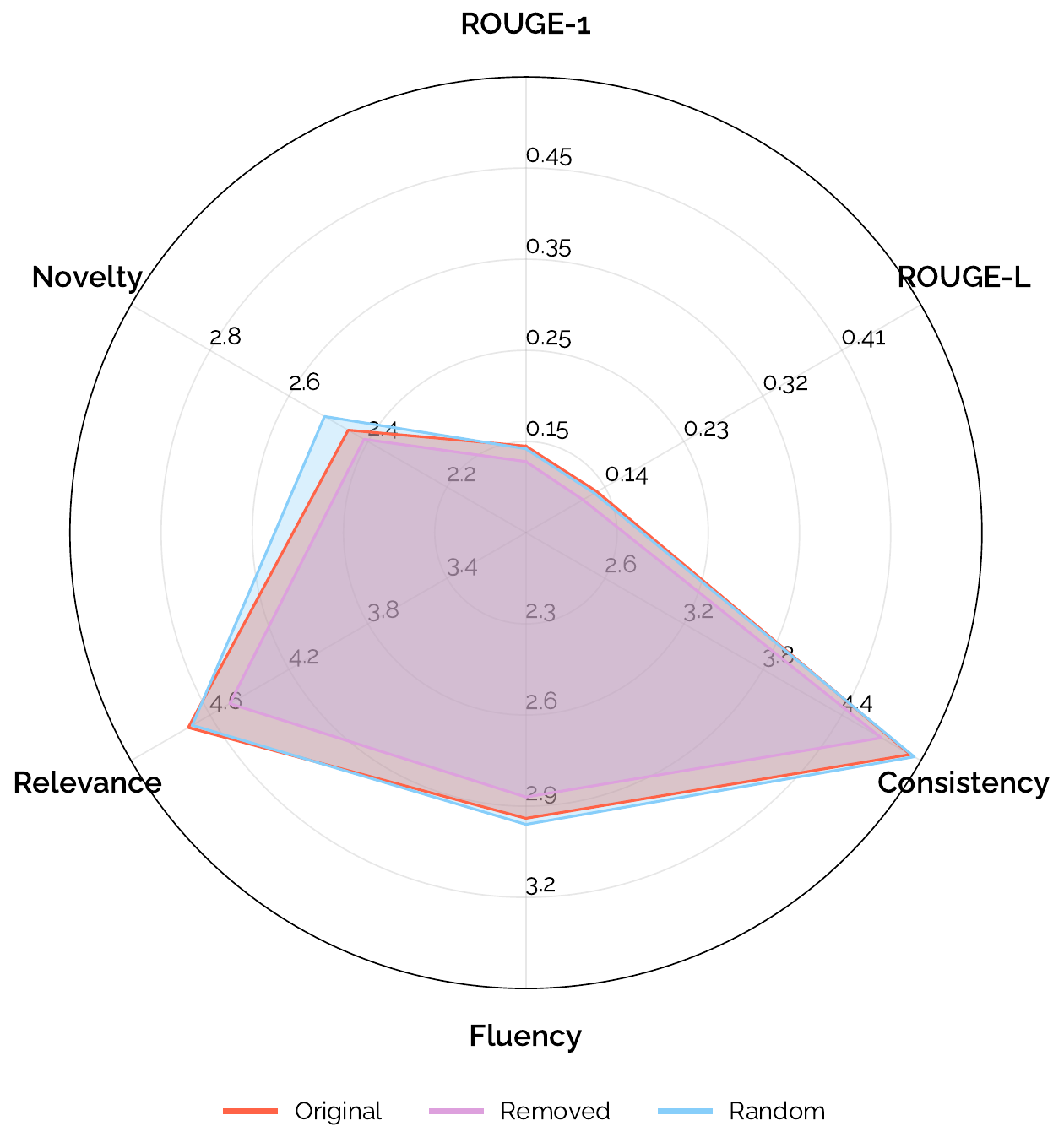}
    \caption{PSW-3}
    \label{fig:psw3-ablation2}
  \end{subfigure}
  \hfill
  \begin{subfigure}[b]{0.45\linewidth}
    \includegraphics[width=\linewidth]{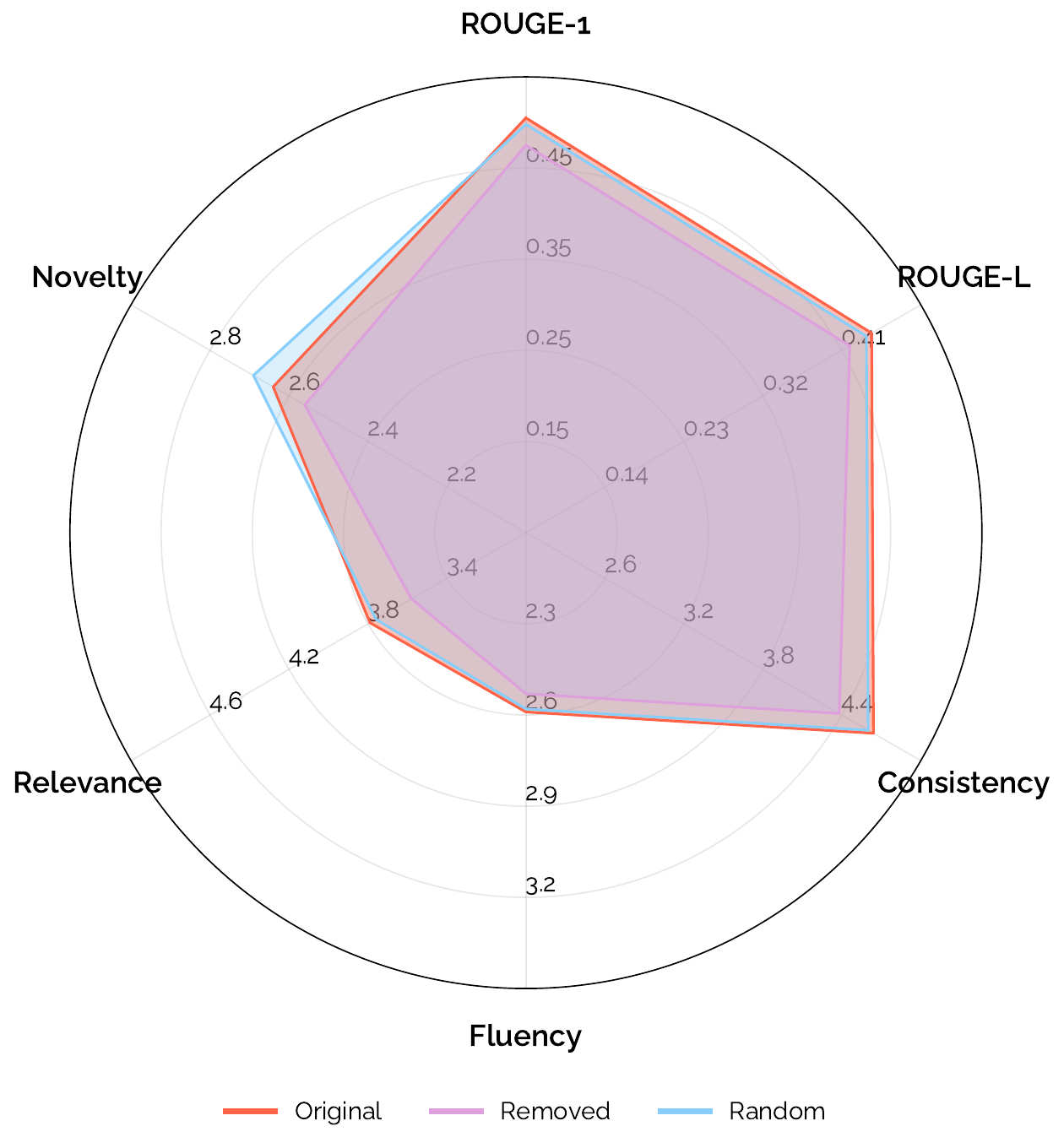}
    \caption{PSW-4}
    \label{fig:psw4-ablation2}
  \end{subfigure}
  
  \caption{\textbf{Impact of user profiling on the performance across three different models:} \textbf{\texttt{Original}}, \textbf{\texttt{Removed}}, and \textbf{\texttt{Random}}. The \textbf{\texttt{Original}} model consistently achieves the highest scores across all datasets.}
  \label{fig:psw_main-ablation2}
\end{figure}

Figures \ref{fig:psw_main}, \ref{fig:psw_main-ablation1}, and \ref{fig:psw_main-ablation2} illustrate the results discussed in Section \ref{sec:main_result} and Appendices \ref{app:ablation_order} and \ref{app:ablation_profile}, respectively.

\clearpage
\section{Details of G-Eval }
\label{sec:prompt_geval}

\begin{table}[tbh!]
\centering

\begin{tabular}{p{0.98\linewidth}}

\hline

\rowcolor{gray!20}{\textbf{Task Description}}\\
\hline

You will be given one result generated for a science paper and several reference papers. Your task is to rate the result using the following criteria.\\
Please make sure you read and understand these instructions carefully. Please keep this document open while reviewing, and refer to it as needed.\\
\hline
\rowcolor{brown!20}{\textbf{Evaluation Criteria}}\\
\hline
\textbf{Consistency (1-5)} -- the factual alignment between the result and the corresponding science paper. A factually consistent result contains only statements entailed by the source document. \\

\textbf{Fluency (1-3)} -- the quality of the result in terms of grammar, spelling, punctuation, word choice, and sentence structure. \\

\textbf{Relevance (1-5)} -- the selection of important content from the source. The result should include only important information from the source document. \\

\textbf{Novelty (1-3)} -- the uniqueness and originality of the result in terms of concept, perspective, and creativity. \\
\hline
\rowcolor{gray!20}{\textbf{Evaluation Task}}\\
\hline
Now, you are working on evaluating this prediction:\\
\{Prediction Text\}\\
Here are some ground truth results for comparison: [result$_1$, result$_2$, \ldots]. \\
\hline
\rowcolor{gray!20}{\textbf{Instruction}}\\
\hline
Please evaluate the prediction using the above criteria. \\
\hline

\end{tabular}
\caption{Prompt template for evaluating the G-Eval metric.}
\label{table:template_novelty}
\end{table}

\newpage

\section{Prompts for LaMP Tasks}
\label{sec:prompt_lamp}

\subsection{User Profile Generation}

\begin{table}[!tbh]
\centering
\begin{tabular}{p{0.98\linewidth}}
\hline
\rowcolor{gray!20}{\textbf{User History}}\\
\hline
I will provide you with some examples of the user's past interactions. Please analyze these to create a user profile. Each example consists of an input and the corresponding output.\\
\hline
\rowcolor{brown!20}
\textbf{Example 1}\\
\textbf{Input:} \{Input Text\}\\
\textbf{Output:} \{Output Text\}\\
\hline
\rowcolor{brown!20}
\textbf{Example 2}\\
\textbf{Input:} \{Input Text\}\\
\textbf{Output:} \{Output Text\}\\
\ldots\\
\hline
\rowcolor{brown!20}
\textbf{Example k}\\
\textbf{Input:} \{Input Text\}\\
\textbf{Output:} \{Output Text\}\\
\hline
\rowcolor{gray!20}{\textbf{Instruction}}\\
\hline
Based on the provided examples, please generate a user profile in the following format:\\
\textbf{Keywords:} [keyword$_1$, keyword$_2$, \ldots]\\
\textbf{Topics:} [topic$_1$, topic$_2$,  \ldots]\\
\textbf{Writing Style:} [style$_1$, style$_2$,  \ldots]\\
\textbf{Preferences:} [preference$_1$, preference$_2$, \ldots]\\
\bottomrule
\end{tabular}
\caption{Prompt template for LaMP User Profile Generation.}
\label{table:template_lamp_profile}
\end{table}

\newpage
\subsection{Personalized Citation Identification (LaMP-1)}

\begin{table}[tbh!]

\centering

\begin{tabular}{p{0.98\linewidth}}

\hline

\rowcolor{gray!20}{\textbf{User Profile}}\\

\hline

Assuming you care a lot about these areas:\\

\textbf{Keywords:} [keyword$_1$, keyword$_2$, keyword$_3$, \ldots]\\

\textbf{Topics:} [topics$_1$, topics$_2$, topics$_3$, \ldots]\\

\hline

\rowcolor{gray!20}{\textbf{User History}}\\

\hline

I give you some titles of papers that you've written. Please imitate your reasons and recommend a paper citation for me. Each example consists of an abstract, the corresponding title, and a description of the writing style and keywords for that title.\\

\hline

\rowcolor{brown!20}

\textbf{Example 1}\\

\textbf{Title:} \{Title Text\}\\

\textbf{Abstract:} \{Abstract Text\}\\

\textbf{Reason:} \{Reason\}\\

\textbf{Citation:} [citation$_1$, citation$_2$, \ldots]\\

\hline

\rowcolor{brown!20}

\textbf{Example 2}\\

\textbf{Title:} \{Title Text\}\\

\textbf{Abstract:} \{Abstract Text\}\\

\textbf{Reason:} \{Reason\}\\

\textbf{Citation:} [citation$_1$, citation$_2$, \ldots]\\

\ldots\\

\hline

\rowcolor{brown!20}

\textbf{Example k}\\

\textbf{Title:} \{Title Text\}\\

\textbf{Abstract:} \{Abstract Text\}\\

\textbf{Reason:} \{Reason\}\\

\textbf{Citation:} [citation$_1$, citation$_2$, \ldots]\\

\hline

\rowcolor{gray!20}{\textbf{Classification Task}}\\

\hline

Now you have written this title:\\

\textbf{Title:} \{Title Text\}\\

\hline

\rowcolor{gray!20}{\textbf{Instruction}}\\

\hline

Please separately analyze the potential relevant connection of \textbf{Reference 1} and \textbf{Reference 2} to this title. You are citing from one of them. Please decide which one it would be:\\

\textbf{Reference 1:} \{option$_1$\}\\

\textbf{Reference 2:} \{option$_2$\}\\

Just answer with [1] or [2] without explanation.\\

\bottomrule

\end{tabular}

\caption{Prompt template for the Personalized Citation Identification (LaMP-1) task.}

\label{table:template_lamp1}

\end{table}

\newpage

\subsection{Personalized News Categorization (LaMP-2) }

\begin{table}[tbh!]

\centering

\begin{tabular}{p{0.98\linewidth}}

\hline

\rowcolor{gray!20}{\textbf{User Profile}}\\

\hline

\hline
Assuming you care a lot about these areas:\\

\textbf{Keywords:} [keyword$_1$, keyword$_2$, keyword$_3$, \ldots]\\
\textbf{Topics:} [topics$_1$, topics$_2$, topics$_3$, \ldots]\\

\hline

\rowcolor{gray!20}{\textbf{User History}}\\

\hline

I give you some titles and articles that you've written with category. Please imitate your reasons for giving this category. Each example consists of an abstract, the corresponding title, and a category of it.\\

\hline

\rowcolor{brown!20}

\textbf{Example 1}\\

\textbf{Article:} \{Article Text\}\\

\textbf{Title:} \{Title Text\}\\

\textbf{Reason:} \{Reason\}\\

\textbf{Category:} [category$_1$, category$_2$, \ldots]\\

\hline

\rowcolor{brown!20}

\textbf{Example 2}\\

\textbf{Article:} \{Article Text\}\\

\textbf{Title:} \{Title Text\}\\

\textbf{Reason:} \{Reason\}\\

\textbf{Category:} [category$_1$, category$_2$, \ldots]\\

\ldots\\

\hline

\rowcolor{brown!20}

\textbf{Example k}\\

\textbf{Article:} \{Article Text\}\\

\textbf{Title:} \{Title Text\}\\

\textbf{Reason:} \{Reason\}\\

\textbf{Category:} [category$_1$, category$_2$, \ldots]\\

\hline

\rowcolor{gray!20}{\textbf{Classification Task}}\\

\hline

Now you have written this article with the title:\\

\textbf{Article:} \{Article Text\}\\

\textbf{Title:} \{Title Text\}\\

\hline

\rowcolor{gray!20}{\textbf{Instruction}}\\

\hline

Which category does this article relate to among the following categories?\\

\textbf{Category 1:} \{option$_1$\}\\
\textbf{Category 2:} \{option$_2$\}\\
\ldots\\
\textbf{Category K:} \{option$_{N}$\}\\

Just answer with the category name without further explanation.\\

\bottomrule

\end{tabular}

\caption{Prompt template for the Personalized News Categorization (LaMP-2) task.}

\label{table:template_lamp2}

\end{table}
\newpage

\subsection{Personalized Product Rating (LaMP-3) }

\begin{table}[tbh!]
\centering
\begin{tabular}{p{0.98\linewidth}}
\hline
\rowcolor{gray!20}{\textbf{User Profile}}\\
\hline
Assuming you have written product reviews with the following characteristics:\\
\textbf{Most Common Rating:} \{score$_\text{most}$\}\\
\textbf{Rating Patterns:} [pattern$_1$, pattern$_2$, \ldots]\\
\hline
\rowcolor{gray!20}{\textbf{User History}}\\
\hline
I provide you with some product reviews you've written, along with their corresponding ratings. Please imitate your reasoning for assigning these ratings. Each example consists of a product review and its rating.\\
\hline
\rowcolor{brown!20}
\textbf{Example 1}\\
\textbf{Product Review:} \{Review Text\}\\
\textbf{Rating:} \{Rating\}\\
\hline
\rowcolor{brown!20}
\textbf{Example 2}\\
\textbf{Product Review:} \{Review Text\}\\
\textbf{Rating:} \{Rating\}\\
\ldots\\
\hline
\rowcolor{brown!20}
\textbf{Example k}\\
\textbf{Product Review:} \{Review Text\}\\
\textbf{Rating:} \{Rating\}\\
\hline
\rowcolor{gray!20}{\textbf{Rating Task}}\\
\hline
Now you have written this new product review:\\
\textbf{Product Review:} \{Review Text\}\\
Based on the review, please analyze its sentiment and how much you like the product. \\

\hline
\rowcolor{gray!20,}{\textbf{Instruction}}\\
\hline

Follow your previous rating habits and these instructions:\\

\hline
\rowcolor{gray!10}
\begin{itemize}[leftmargin=*, itemsep=1pt,topsep=-5pt,parsep=0pt]
\item If you feel satisfied with this product or have concerns but it's good overall, it should be rated 5.
\item If you feel good about this product but notice some issues, it should be rated as 4.
\item If you feel OK but have concerns, it should be rated as 3.
\item If you feel unsatisfied with this product but it's acceptable for some reason, it should be rated as 2.
\item If you feel completely disappointed or upset, it should be rated 1.
\end{itemize}\\

\hline

Your most common rating is \{score$_\text{most}$\}. You must follow this rating pattern faithfully and answer with the rating without further explanation.\\
\bottomrule
\end{tabular}
\label{table:template_lamp3u}
\caption{Prompt template for the Personalized Product Review Rating (LaMP-3) task.}
\end{table}
\newpage

\subsection{Personalized News Headline Generation (LaMP-4) }

\begin{table}[tbh!]
\centering
\begin{tabular}{p{0.98\linewidth}}
\hline
\rowcolor{gray!20}{\textbf{User Profile}}\\
\hline
Assuming you have written headlines with the following characteristics:\\
\textbf{Writing Style:} [style$_1$, style$_2$, \ldots]\\
\textbf{Content Patterns:} [patterns$_1$, patterns$_2$, \ldots]\\
\hline
\rowcolor{gray!20}{\textbf{User History}}\\
\hline
I will provide you with some news articles along with the headlines you've written for them. Please imitate your writing style and content patterns when generating a new headline. Each example consists of a news article and its corresponding headline.\\
\hline
\rowcolor{brown!20}
\textbf{Example 1}\\
\textbf{Article:} \{Article Text\}\\
\textbf{Headline:} \{Headline\}\\
\hline
\rowcolor{brown!20}
\textbf{Example 2}\\
\textbf{Article:} \{Article Text\}\\
\textbf{Headline:} \{Headline\}\\
\ldots\\
\hline
\rowcolor{brown!20}
\textbf{Example k}\\
\textbf{Article:} \{Article Text\}\\
\textbf{Headline:} \{Headline\}\\

\hline
\rowcolor{gray!20}{\textbf{Generation Task}}\\
\hline

Now that you have been given this news article:\\
\textbf{Article:} \{Article Text\}\\

\hline
\rowcolor{gray!20}{\textbf{Instruction}}\\
\hline

Please write a headline following your previous writing styles and habits. If you have written headlines with similar content, you could reuse those headlines and mimic their content.\\

\bottomrule
\end{tabular}
\caption{Prompt template for the Personalized News Headline Generation (LaMP-4) task.}
\label{table:template_lamp4u}
\end{table}

\newpage

\subsection{Personalized Scholarly Title Generation (LaMP-5) }

\begin{table}[tbh!]
\centering
\begin{tabular}{p{0.98\linewidth}}
\hline
\rowcolor{gray!20}{\textbf{User Profile}}\\
\hline
Assuming you have written scholarly titles with the following characteristics:\\
\textbf{Writing Style:} [style$_1$, style$_2$, \ldots]\\
\textbf{Title Patterns:} [pattern$_1$, pattern$_2$, \ldots]\\
\hline
\rowcolor{gray!20}{\textbf{User History}}\\
\hline
I will provide you with some research paper abstracts along with the titles you've written for them. Please imitate your writing style and title patterns when generating a new title. Each example consists of a paper abstract and its corresponding title.\\
\hline
\rowcolor{brown!20}
\textbf{Example 1}\\
\textbf{Abstract:} \{Abstract Text\}\\
\textbf{Title:} \{Title\}\\
\hline
\rowcolor{brown!20}
\textbf{Example 2}\\
\textbf{Abstract:} \{Abstract Text\}\\
\textbf{Title:} \{Title\}\\
\ldots\\
\hline
\rowcolor{brown!20}
\textbf{Example k}\\
\textbf{Abstract:} \{Abstract Text\}\\
\textbf{Title:} \{Title\}\\

\hline
\rowcolor{gray!20}{\textbf{Generation Task}}\\
\hline

Now that you have been given this paper abstract:\\
\textbf{Abstract:} \{Abstract Text\}\\
\hline
\rowcolor{gray!20}{\textbf{Instruction}}\\
\hline

Please write a title following your previous style and habits, keeping it clear, accurate, and concise. \\
\bottomrule
\end{tabular}
\caption{Prompt template for the Personalized Scholarly Title Generation (LaMP-5) task.}
\label{table:template_lamp5u}
\end{table}

\newpage

\subsection{Personalized Email Subject Generation (LaMP-6) }

\begin{table}[tbh!]
\centering
\begin{tabular}{p{0.98\linewidth}}
\hline
\rowcolor{gray!20}{\textbf{User Profile}}\\
\hline
Assuming you care a lot about these areas:\\
\textbf{Keywords:} [keyword$_1$, keyword$_2$, keyword$_3$, \ldots]\\
\textbf{Topics:} [topics$_1$, topics$_2$, topics$_3$, \ldots]\\
\hline
\rowcolor{gray!20}{\textbf{User History}}\\
\hline
Let's say there are some emails you've written. Please mimic the style of these examples.
Each example consists of email content, the corresponding subject, and a description
of the writing style for that title.\\
\hline
\rowcolor{brown!20}
\textbf{Example 1}\\
\textbf{Content:} \{Email Content\}\\
\textbf{Writing Style:} \{Style\}\\
\textbf{Subject:} \{Email Subject\}\\
\hline
\rowcolor{brown!20}
\textbf{Example 2}\\
\textbf{Content:} \{Email Content\}\\
\textbf{Writing Style:} \{Style\}\\
\textbf{Subject:} \{Email Subject\}\\
\ldots \\
\hline
\rowcolor{brown!20}
\textbf{Example k}\\
\textbf{Content}: \{Email Content\}\\
\textbf{Writing Style:} \{Style\}\\
\textbf{Subject}: \{Email Subject\}\\
\hline
\rowcolor{gray!20}
\textbf{Generation Task}\\
\hline
Now that you have been given this email content:\\
\textbf{Content:} \{Email Content\}\\
\hline
\rowcolor{gray!20}
\textbf{Instruction}\\
\hline
Write a title following your previous style and habits. Just answer with the subject without further explanation. \\

\bottomrule
\end{tabular}
\caption{Prompt template for the Personalized Email Subject Generation (LaMP-6) task.}
\label{table:template_llm}
\end{table}

\newpage
\subsection{Personalized Tweet Paraphrasing (LaMP-7) }

\begin{table}[tbh!]
\centering

\begin{tabular}{p{0.98\linewidth}}
\hline
\rowcolor{gray!20}{\textbf{User Profile}}\\
\hline
Assuming you have written tweets with the following characteristics:\\
\textbf{Writing Style:} [style$_1$, style$_2$, \ldots]\\
\textbf{Tone:} [tone$_1$, tone$_2$, \ldots]\\
\textbf{Length:} [length$_1$, length$_2$, \ldots]\\
\hline
\rowcolor{gray!20}{\textbf{User History}}\\
\hline
I will provide you with some original tweets along with the paraphrased versions you've written for them. When paraphrasing a new tweet, please imitate your writing style, tone, and typical length. Each example consists of an original tweet and its paraphrased version.\\
\hline
\rowcolor{brown!20}
\textbf{Example 1}\\
\textbf{Original Tweet:} \{Tweet Text\}\\
\textbf{Paraphrased Tweet:} \{Paraphrased Text\}\\
\hline
\rowcolor{brown!20}
\textbf{Example 2}\\
\textbf{Original Tweet:} \{Tweet Text\}\\
\textbf{Paraphrased Tweet:} \{Paraphrased Text\}\\
\ldots\\
\hline
\rowcolor{brown!20}
\textbf{Example k}\\
\textbf{Original Tweet:} \{Tweet Text\}\\
\textbf{Paraphrased Tweet:} \{Paraphrased Text\}\\
\hline
\rowcolor{gray!20}{\textbf{Generation Task}}\\
\hline
Now that you have been given this tweet:\\
\textbf{Original Tweet:} \{Tweet Text\}\\
\hline
\rowcolor{gray!20}{\textbf{Instruction}}\\
\hline
Please paraphrase it with the following instructions:
\begin{itemize}[leftmargin=*, itemsep=0pt, topsep=0pt]
\item You must use tweet styles and tones.
\item You must keep it faithful to the given tweet with similar keywords and length.
\end{itemize} \\

\bottomrule
\end{tabular}
\caption{Prompt template for the Personalized Tweet Paraphrasing (LaMP-7) task.}
\label{table:template_lamp7u}
\end{table}

\newpage
\section{Prompts for PSW Tasks}
\label{sec:prompt_psw}
\subsection{Research Interests Generation (UP-0)  }

\begin{table}[!tbh]
\centering
\begin{tabular}{p{0.98\linewidth}} 
\hline
\rowcolor{gray!20}{\textbf{User History}}\\
\hline
I will provide you with some research papers you've authored. Please summarize your top research interests based on these papers. Each paper consists of a title and abstract.\\
\hline
\rowcolor{brown!20}
\textbf{Paper 1}\\  
\textbf{Title:} \{Title Text\}\\
\textbf{Abstract:} \{Abstract Text\}\\
\hline
\rowcolor{brown!20}
\textbf{Paper 2}\\ 
\textbf{Title:} \{Title Text\}\\
\textbf{Abstract:} \{Abstract Text\}\\
\ldots\\
\hline
\rowcolor{brown!20}  
\textbf{Paper k}\\
\textbf{Title:} \{Title Text\}\\  
\textbf{Abstract:} \{Abstract Text\}\\
\hline
\rowcolor{gray!20}{\textbf{Instruction}}\\
\hline
Please summarize your top three research interests based on the provided papers in the following format:\\
\textbf{Research Interests:} [interest$_1$, interest$_2$, interest$_3$, \ldots] \\
\bottomrule
\end{tabular}
\caption{Prompt template for the Research Interests Generation (UP-0) task.}
\label{table:template_up0}
\end{table}

\newpage
\subsection{Personalized Research Paper Title Generation (PSW-1) }

\begin{table}[!tbh]
\centering
\begin{tabular}{p{0.98\linewidth}}
\hline
\rowcolor{gray!20}{\textbf{User Profile}}\\
\hline
Assuming you are an expert researcher with the following research interests:\\
\textbf{Research Interests:} [interest$_1$, interest$_2$, interest$_3$, \ldots]\\
\hline
\rowcolor{gray!20}{\textbf{User History}}\\
\hline
Here are some titles and abstracts from papers you have authored:\\
\hline
\rowcolor{brown!20}
\textbf{Paper 1}\\
\textbf{Title:} \{Title\}\\
\textbf{Abstract:} \{Abstract\}\\
\hline
\rowcolor{brown!20}
\textbf{Paper 2}\\
\textbf{Title:} \{Title\}\\
\textbf{Abstract:} \{Abstract\}\\
\ldots\\
\hline
\rowcolor{brown!20}
\textbf{Paper k}\\
\textbf{Title:} \{Title\}\\
\textbf{Abstract:} \{Abstract\}\\
\hline
\rowcolor{gray!20}{\textbf{Brainstorm Task}}\\
\hline
Here are some related papers for reference, each with a title:\\
\textbf{Reference 1:} \{Title\}\\
\textbf{Reference 2:} \{Title\}\\
\ldots\\
\textbf{Reference N:} \{Title\}\\
\hline
\rowcolor{gray!20}{\textbf{Instruction}}\\
\hline
Considering your research interests, previous works, and reference papers, please brainstorm the most promising title for your new research paper.\\
\bottomrule
\end{tabular}
\caption{Prompt template for the Personalized Research Paper Title Generation (PSW-1) task.}
\label{table:template_psw1}
\end{table}

\newpage

\subsection{Research Question Generation (PSW-2)  }

\begin{table}[!tbh]
\centering
\begin{tabular}{p{0.98\linewidth}}
\hline
\rowcolor{gray!20}{\textbf{User Profile}}\\
\hline
Assuming you are an expert researcher with the following research interests:\\
\textbf{Research Interests:} [interest$_1$, interest$_2$, interest$_3$, \ldots]\\
\hline
\rowcolor{gray!20}{\textbf{User History}}\\  
\hline
Here are some titles and abstracts from papers you have authored:\\
\hline
\rowcolor{brown!20}
\textbf{Paper 1}\\
\textbf{Title:} \{Title\}\\
\textbf{Abstract:} \{Abstract\}\\  
\hline
\rowcolor{brown!20}
\textbf{Paper 2}\\
\textbf{Title:} \{Title\}\\
\textbf{Abstract:} \{Abstract\}\\
...\\
\hline  
\rowcolor{brown!20}
\textbf{Paper k}\\
\textbf{Title:} \{Title\}\\
\textbf{Abstract:} \{Abstract\}\\
\hline
\rowcolor{gray!20}{\textbf{Brainstorm Task}}\\
\hline  
Now you are working on a new paper with the following title:\\
\textbf{Title:} \{Title\}\\
\hline
\rowcolor{gray!20}{\textbf{Instruction}}\\
\hline  
Considering the title and research background, please propose the top 3 research questions you aim to address in this new paper. \\
\bottomrule
\end{tabular}
\caption{Prompt template for the Research Question Generation (PSW-2) task.}  
\label{table:template_psw2}
\end{table}

\newpage

\subsection{Paper Abstract Generation (PSW-3)  }

\begin{table}[!tbh]
\centering
\begin{tabular}{p{0.98\linewidth}}
\hline
\rowcolor{gray!20}{\textbf{User Profile}}\\
\hline
Assuming you are an expert researcher with the following research interests:\\  
\textbf{Research Interests:} [interest$_1$, interest$_2$, interest$_3$, ...]\\
\hline
\rowcolor{gray!20}{\textbf{User History}}\\
\hline
Here are some titles and abstracts from papers you have authored:\\  
\hline
\rowcolor{brown!20} 
\textbf{Paper 1}\\
\textbf{Title:} \{Title\}\\
\textbf{Abstract:} \{Abstract\}\\ 
\hline
\rowcolor{brown!20}
\textbf{Paper 2}\\  
\textbf{Title:} \{Title\}\\
\textbf{Abstract:} \{Abstract\}\\
...\\
\hline
\rowcolor{brown!20}  
\textbf{Paper k}\\
\textbf{Title:} \{Title\}\\
\textbf{Abstract:} \{Abstract\}\\  
\hline
\rowcolor{gray!20}{\textbf{Generation Task}}\\
\hline
Now you are working on a new paper with the following title: \\
\textbf{Title:} \{Title\} \\
And you are focusing on solving the following research questions: [question$_1$, question$_2$, \ldots] \\
\hline
\rowcolor{gray!20}{\textbf{Instruction}}\\
\hline

Considering the title, research questions, and your writing style in previous abstracts, please write an abstract for this new paper. \\
\bottomrule
\end{tabular}
\caption{Prompt template for the Paper Abstract Generation (PSW-3) task.}
\label{table:template_psw3}  
\end{table}

\newpage

\subsection{Paper Title Generation (PSW-4) }

\begin{table}[!tbh]
\centering  
\begin{tabular}{p{0.98\linewidth}}
\hline
\rowcolor{gray!20}{\textbf{User Profile}}\\  
\hline
Assuming you are an expert researcher with the following research interests:\\
\textbf{Research Interests:} [interest$_1$, interest$_2$, interest$_3$, ...]\\  
\hline
\rowcolor{gray!20}{\textbf{User History}}\\
\hline  
Here are some titles and abstracts from papers you have authored:\\
\hline
\rowcolor{brown!20}   
\textbf{Paper 1}\\
\textbf{Title:} \{Title\}\\
\textbf{Abstract:} \{Abstract\}\\   
\hline
\rowcolor{brown!20}
\textbf{Paper 2}\\   
\textbf{Title:} \{Title\}\\
\textbf{Abstract:} \{Abstract\}\\
...\\
\hline   
\rowcolor{brown!20}
\textbf{Paper k}\\
\textbf{Title:} \{Title\}\\   
\textbf{Abstract:} \{Abstract\}\\
\hline
\rowcolor{gray!20}{\textbf{Generation Task}}\\   
\hline
Now, you are working on a new paper with the following abstract:\\ 
\textbf{Abstract:} \{Abstract\}\\ 
And you are focusing on solving the following research questions: [question$_1$, question$_2$, \ldots] \\
\hline
\rowcolor{gray!20}{\textbf{Instruction}}\\   
\hline
Considering the abstract and your title writing style in previous papers, please generate a title for this new paper. The title should be clear and concise and reflect the main topic of the abstract as well as your research questions. \\
\bottomrule
\end{tabular}
\caption{Prompt template for the Paper Title Generation (PSW-4) task.}
\label{table:template_psw4}   
\end{table}

\end{document}